\documentclass[journal]{IEEEtran}

\usepackage{url}

\usepackage[hidelinks]{hyperref}

\usepackage{color}

\usepackage{flushend}

\usepackage{cite}
\usepackage{amssymb}

\ifCLASSINFOpdf
   \usepackage[pdftex]{graphicx}
\else
   \usepackage[dvips]{graphicx}
\fi
\usepackage{graphbox}
\usepackage{subfigure}
\usepackage{multirow}
\usepackage{float}

\usepackage[cmex10]{amsmath}
\DeclareMathOperator*{\argmax}{arg\,max}

\usepackage{array}

\usepackage{balance}

\usepackage{algorithm}
\usepackage{algorithmic}

\usepackage{tabularx}

\begin{document}

\title{Intelligent Trainer for Model-Based Deep Reinforcement Learning}
\author{Yuanlong~Li,~\IEEEmembership{Member,~IEEE,}
        Linsen~Dong,~\IEEEmembership{Student Member,~IEEE,}
        Xin~Zhou,~\IEEEmembership{Member,~IEEE,}
        Yonggang~Wen,~\IEEEmembership{Senior Member,~IEEE,}
        and Kyle~Guan~\IEEEmembership{Member,~IEEE}
\thanks {Manuscript received... This work was supported in part by EIRP02 Grant from Singapore EMA, GDCR01 Grant from Singapore IMDA.}
\thanks{Yuanlong Li, Xin Zhou, Yonggang Wen and  Linsen Dong are with School of Computer Science and Engineering, Nanyang Technological University, Nanyang Avenue, Singapore 639798. Email: \{liyuanl, ygwen\}@ntu.edu.sg, LINSEN001@e.ntu.edu.sg}
\thanks{Kyle Guan is with Bell Labs, Nokia. Email: kyle.guan@nokia.com. }
}

\maketitle

\begin{abstract}
Model-based reinforcement learning (MBRL) has been proposed as a promising alternative solution to tackle the high sampling cost challenge in the canonical reinforcement learning (RL), by leveraging a learned model to generate synthesized data for policy training purpose. The MBRL framework, nevertheless, is inherently limited by the convoluted process of jointly learning control policy and configuring hyper-parameters (e.g., global/local models, real and synthesized data, etc). The training process could be tedious and prohibitively costly. In this research, we propose an ``reinforcement on reinforcement'' (RoR) architecture to decompose the convoluted tasks into two layers of reinforcement learning. The inner layer is the canonical model-based RL training process environment (TPE), which learns the control policy for the underlying system and exposes interfaces to access states, actions and rewards. The outer layer presents an RL agent, called as AI trainer, to learn an optimal hyper-parameter configuration for the inner TPE. This decomposition approach provides a desirable flexibility to implement different trainer designs, called as ``train the trainer''. In our research, we propose and optimize two alternative trainer designs: 1) a uni-head trainer and 2) a multi-head trainer. Our proposed RoR framework is evaluated for five tasks in the OpenAI gym (i.e., Pendulum, Mountain Car, Reacher, Half Cheetah and Swimmer). Compared to three  other baseline algorithms, our proposed Train-the-Trainer algorithm has a competitive performance in auto-tuning capability, with upto 56\% expected sampling cost saving without knowing the best parameter setting in advance. The proposed trainer framework can be easily extended to other cases in which the hyper-parameter tuning is costly.
\end{abstract}

\section{Introduction}

\IEEEPARstart{R}einforcement learning (RL) \cite{sutton2018reinforcement}, owing to its flexibility derived from data-driven nature, has recently regained tremendous momentum in research and industry applications. RL, in comparison to supervised and unsupervised learning, addresses how intelligent agents should take actions in an environment, aiming to maximize a chosen cumulative reward function. For example, an RL agent controlling a robot arm to grab an object will observe the current state of the arm, issue an action to the arm and after the action being taken, collect the reward, signifying whether the object has been grabbed or not, and new state information to train its policy. This interaction between the agent and the environment allows for progressively learning the best control policy in complex systems. Previously RL has been adopted to solve problems like robot arm control, maze solving and game playing, reducing the human intervention in system modeling. Recently, RL, in combination with the emerging deep learning techniques (so called the deep reinforcement learning - DRL) \cite{mnih2013playing}, has become a popular choice for large complex system control. This trend started with the huge success of AlphaGO. At the same time, researchers have also made breakthrough in complex system control in continuous domains, via DRL algorithms, for example, Deep Deterministic Policy Gradients (DDPG)\cite{lillicrap2015continuous} and Trust Region Policy Optimization (TRPO)\cite{schulman2015trust}. As a result, the transformative nature of RL and DRL have been driving its regained popularity in both academia and industry.

However, many practical industry applications face great challenges in adopting RL solutions, especially when it is costly to acquire data for policy training purpose. On the demand side, the performance of RL algorithms in general hinges upon a huge amount of operational data to train the control policy. On the supply side, acquiring a huge amount of training data from operational systems might be prohibitively costly, in resource usage and/or time consumption. For instance, in a robotic control problem~\cite{ICLR16-hausknecht}, the DRL agent can learn to score a goal with high probability, only after about three million samples observed. As result, training with a real robot to do the task in this case may take millions seconds, rendering the system unacceptable in most application scenarios.

To tackle this challenge with training data, researchers have previously proposed a model-based reinforcement learning (MBRL)~\cite{sutton1991dyna}. In the MBRL framework, the data collected from the real-world systems are used to train a system dynamic model, which is in turn used to generate synthesized data for policy training. The generated data, together with the real world data, are used to train the target controller and search for sensible actions to maximize the accumulative reward. Generally, producing synthesized data in a cyber environment is relatively inexpensive, as such the MBRL has the advantage of low data sampling cost. This comparative advantage contributes to the fact that the MBRL has been a popular approach in robot arm training \cite{deisenroth2011learning} and online tree search based planning \cite{guo2014deep,weber2017imagination,pascanu2017learning}.

\begin{figure}[]
\centering
\subfigure[]{
	\raisebox{0.5\height}{\includegraphics[width=0.45\columnwidth]{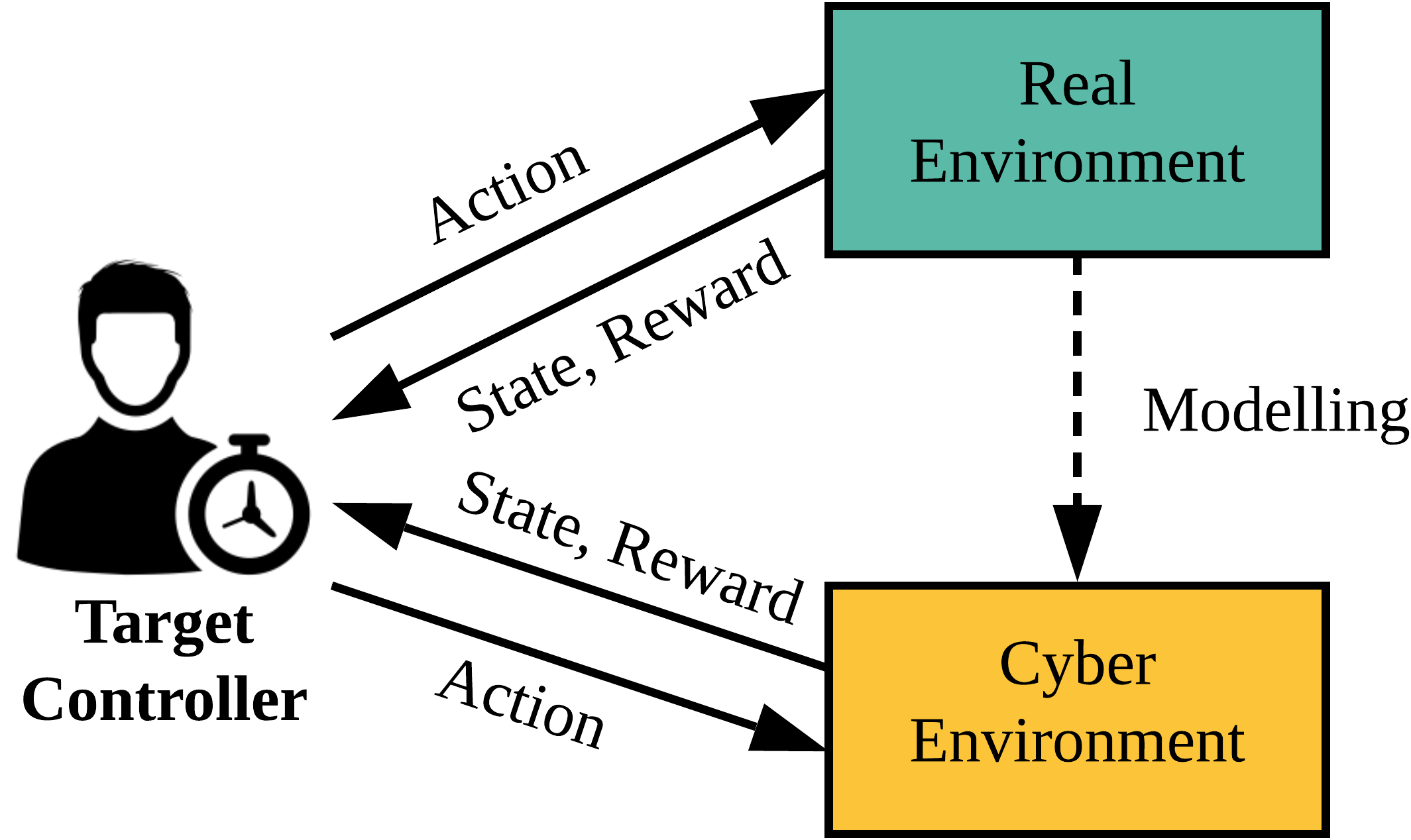}}
}
\subfigure[]{
	\includegraphics[width=0.45\columnwidth]{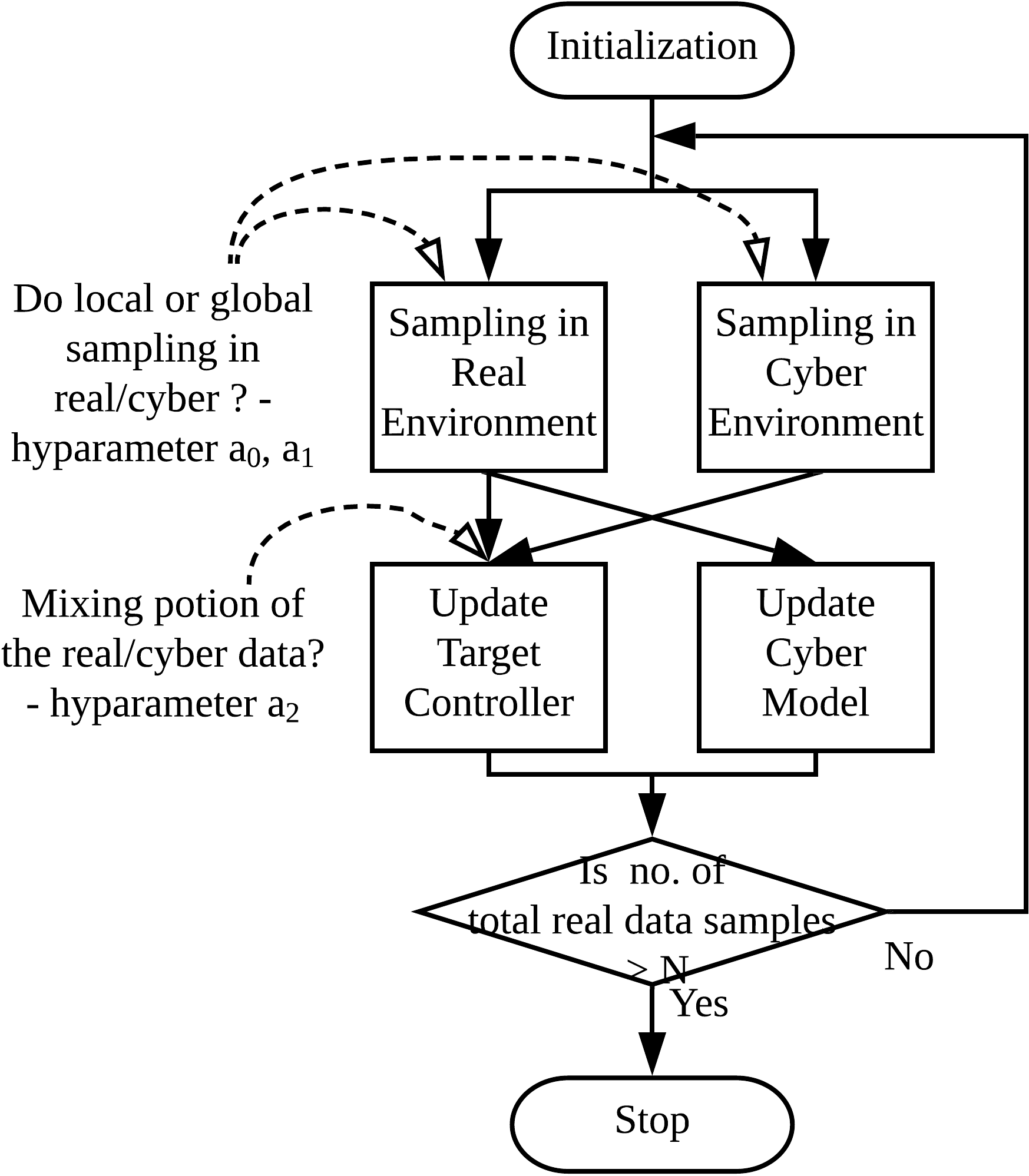}
}
\caption{Illustration of MBRL algorithm using the model as a data source: (a) The data flow of MBRL, where the cyber environment is used to generate synthetic training data for the target controller. (b) Typical training flow of MBRL, in which we indicate the settings that are usually manually set.}
\label{fig:mbrl_flow}
\end{figure}

In real applications, the adoption of the MBRL framework is limited by the manual configuration of some crucial parameters. As illustrated in Figure \ref{fig:mbrl_flow}, the data acquired from the physical environment would be used for two purposes, namely:
\begin{itemize}
	\item System model generation. The model is trained to mimic the real system in that, given the current state and action to take, it predicts the next system state. The learned model can be trained/used in a global or local \cite{weber2017imagination} manner. The global manner means that the model is trained or utilized to generate data samples from the whole state space and can favor global exploration. The local manner is to train or utilize the model to generate data samples in certain constrained subspace, and thus can reinforce local exploitation. In Fig. \ref{fig:mbrl_flow}, parameters $a_0$ and $a_1$ control whether to go global or local in the training and sampling procedure of the system model.
	\item Control policy training. The collected data from physical environment are also used in the training of the target controller, together with the cyber data generated from the learned model. In this case, the portion of the cyber data to use requires proper configuration to achieve the desired outcome. As to be shown in the experimental results of this paper, the proper setting can vary from case to case: for certain cases using more cyber data can be helpful; while for other cases it may lead to serious performance degeneration. In Fig. \ref{fig:mbrl_flow} parameter $a_2$ controls this setting.
\end{itemize}
We refer these configuration parameters introduced by the model as model-related hyper-parameter setting. In previous research, these parameters are manually tried in training stage, often resulting in additional time and/or resource cost\footnote{A naive approach to potentially solve this problem is to re-train the controller with different parameter settings with the collected data samples in the first trial. Such solution will not incur additional sampling cost. However, the ''supervised" learning approach may not work well for the RL case, as the training performance of such a policy is largely determined by the data used in training. If the data used in training are sampled by an under-performed policy, they may lack of important samples that can lead to better performance, making the re-training useless.}. It follows that an autoML solution for MBRL is highly demanded.

\begin{figure}[]
\centering
\includegraphics[width=0.7\columnwidth] {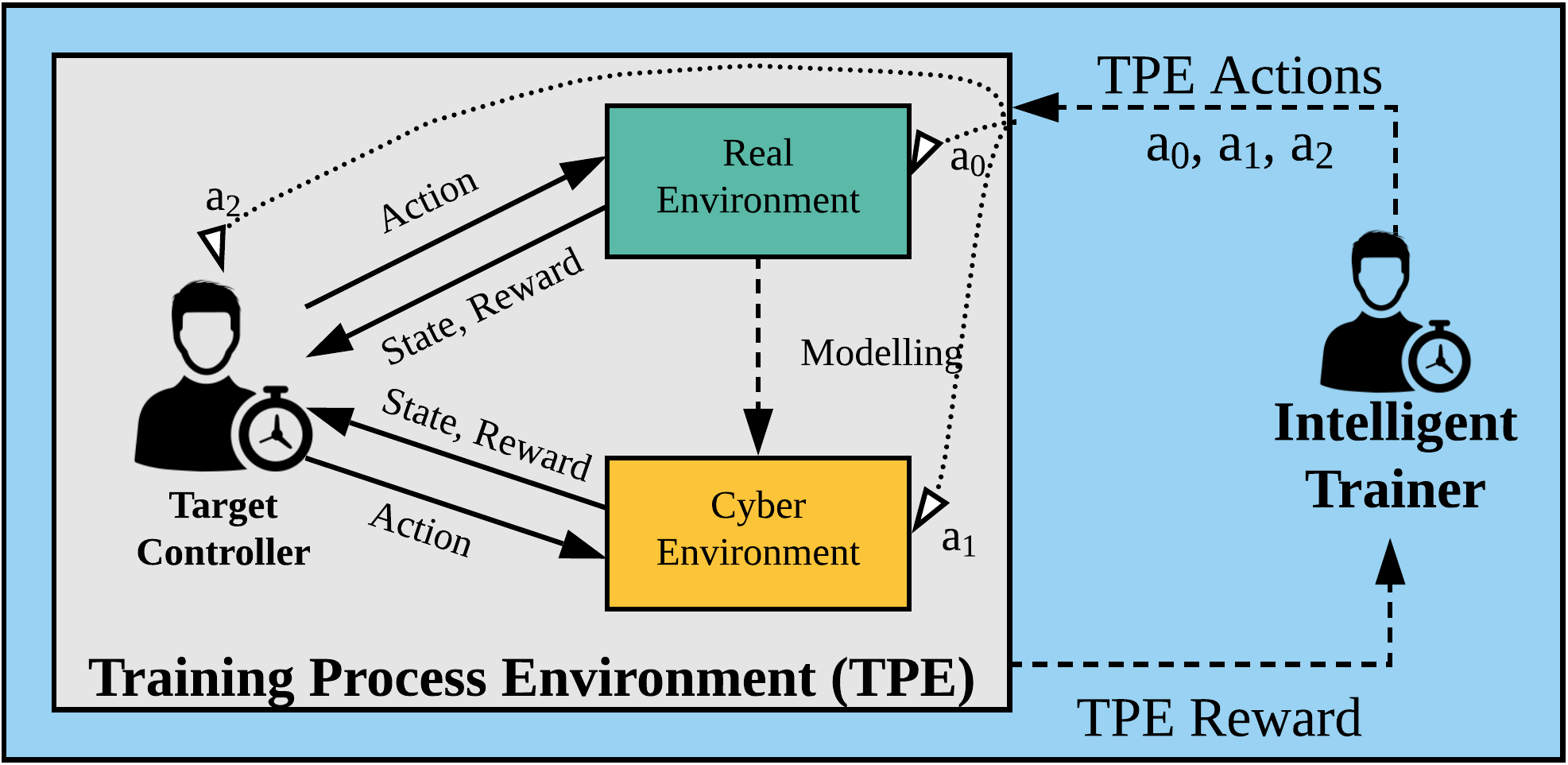}
\caption{Illustration of ``Reinforcement on Reinforcement'' (RoR) framework. The inner box encapsulates a standard MBRL into a training processing environment (TPE) as the inner layer. In the outer box we introduce an intelligent trainer as the outer layer, controlling the optimization of the MBRL training in the TPE.}
\label{fig:RoR}
\end{figure}

In this research, we propose an autoML solution for the MBRL framework, aiming to tackle the hyper-parameter setting challenge. Our proposed solution adopts a ``reinforcement on reinforcement'' (RoR) design architecture to learn the optimal model-related parameters and training/sampling settings in an online manner. Specifically, as illustrated in Fig. \ref{fig:RoR}, we first encapsulate the canonical model-based RL training process into a standard RL environment (Training Process Environment, TPE in Fig. \ref{fig:RoR}) with exposed state, action and reward interfaces, as the inner RL layer. In the outer layer, we introduce an intelligent trainer, as an RL agent, to interact with the inner layer, and control the sampling and training process of the target controller in TPE. Such layered architecture is embedded with an inherent decomposition between the training process and the controlling of the training process in the inner layer,  greatly liberating its applicability in more generalized RL scenarios. In comparison with the existing approaches that directly modify the training algorithm of the target controller \cite{pascanu2017learning}, our design can work with different MBRL controllers and with different trainer designs. We call the latter as ``train the trainer'' design.

Our research intends to optimize the ``train the trainer'' design for better learning efficiency in the outer layer of the RoR architecture, and validate our design over widely-accepted benchmark cases in the openAI gym. First, we propose two alternative trainer designs:
\begin{itemize}
	\item Uni-head trainer. This approach is to implement a single trainer, cast into a DQN controller, to learn in an online manner to optimize the sampling and training in the inner layer.
	\item Multi-head trainer. This approach is to implement an ensemble trainer, comprising of multiple trainers that take independent actions in their training processes, and are ranked across themselves to provide a quantitative comparison of their respective actions.
\end{itemize}
We implement both trainer designs in Tensorflow for five benchmark cases (i.e., Pendulum, Mountain Car, Reacher, Half Cheetah, and Swimmer) and evaluate their performance in learning the best control policy under external constraints. Our evaluation is compared against three baseline algorithms, including a model-free RL algorithm, a MBRL algorithm with randomly hyper-parameter settings and a MBRL algorithm with fixed hyper-parameter settings. Our numerical investigations show that our proposed framework outperforms the aforementioned baseline algorithms in overall performance across different test cases supposing the best parameter settings are unknown. Specifically, our proposed framework can achieve the following results:
\begin{itemize}
	\item For the same learned policy quality, our proposed RoR framework can achieve an expected sampling cost saving upto 56\%, over the average cost of the three baseline algorithms.
	\item Given the same sampling budget, our proposed RoR framework can achieve a policy quality on par with the best policy available, without the prior requirement of knowing the best parameter setting across all the benchmark cases.
\end{itemize}
These evaluations suggest that our proposed RoR framework can be readily applied to emerging industrial applications, with cost concerns. For example, in data center room cooling control application, we can use the trainer framework to properly utilize the computational fluid dynamics (CFD) model with the real monitoring data to train an air cooling unit controller \cite{li2017transforming}. At the same time, it can shed new lights on model-based RL research by leveraging the RoR framework for autoML empowerment. Specifically, it can serve as a general framework that can work with different RL training algorithms and could also be a potential solution for other learning tasks in which online adaptive parameter setting is demanded. We have released the open-source code of our proposed RoR framework at \cite{trainer_repo}, for the research community to further develop new applications and algorithms.

The remainder of this paper is organized as follows. Section \ref{stn:approach} provides a detailed description of the proposed trainer framework, including its key components, uni-head trainer design, and ensemble trainer design. Section \ref{stn:experiment} presents the numerical evaluation results of the proposed framework. Section \ref{stn:related} briefly reviews the related works. Section \ref{stn:summary} concludes the whole paper.

\section{RoR: Reinforcement on Reinforcement Architecture}
\label{stn:approach}
The overall architecture of the proposed intelligent trainer framework is shown in Fig. \ref{fig:RoR}. The inner layer, i.e., the Training Process Environment (TPE) is a standard model-based DRL system utilizing the model as a data source to train the target controller. The training data are provided by the physical environment, which represents the real-world system, and the cyber environment, which is an emulator of the physical system. The emulator can be either knowledge-based or learning-based (e.g., a neural network prediction model). The outer layer, i.e., the intelligent trainer, is also an RL agent that controls and optimizes the sampling and training process of the target controller in the real and cyber environment via feedbacks and action outputs. Thus, the proposed framework can be considered as a ``reinforcement on reinforcement'' architecture. Such modularized design can easily work for different kinds of target controller training algorithms (such as DDPG, TRPO) and the extra layer of intelligent trainer can be any optimizer that can output the control action when given a TPE observation.

In the following we first present the inner layer of the proposed trainer framework to introduce how we encapsulate the standard training process of MBRL as an RL environment. We design the TPE with two goals. First, it should be formulated as a standard RL environment, such that any agent can interact with it. Second, the TPE shall expose the action interface to control the sampling and training settings of the MBRL, and the reward information which can encourage an agent to minimize the sampling cost in the real environment to train the target controller to a target performance. To achieve these goals, the TPE is designed with two major functions and three RL interfaces as following.

\subsection{Basic Functions of TPE}
The TPE has two major functions that can be executed to complete the entire training process of a general MBRL:
\begin{itemize}
 \item \textit{Initialization}: execute initialization tasks for the MBRL training process. These tasks include initializing the real training environment, the cyber emulator, and the target controller.
 \item \textit{Step(state, action)}: execute one step of training of the MBRL algorithm. This process includes sampling from the real and cyber environment, training the target controller, and training the dynamic model of cyber emulator. Note that in each step, we keep the number of real data samples to sample fixed ($K_r$) while optimize the amount $K_c$ of cyber data used in the training. We found that such design is more stable in implementation as it can provide a tractable evaluation of the policy by measuring the received reward from the real environment. With such setting, the TPE exposes action interfaces to determine how many cyber data to use and how to do the sampling, and reward information to encourage a trainer agent to train the target controller to a better performance.
\end{itemize}

With the TPE, MBRL training process can be executed by calling repeatedly calling the Step function after the Initialization. The detailed training algorithm used to train the target controller will be embedded in the Step function.

\subsection{RL Elements of TPE}
For the interaction between TPE and the intelligent trainer, we define three interfaces \textit{State}, \textit{Action}, and \textit{Reward} of TPE as follows. To distinguish the RL components in different layers, in the following, superscript $\xi$ is used to indicate variables in the target controller layer, while $\Xi$ is used to indicate variables in the intelligent trainer layer.
\begin{itemize}
 \item \textit{State}: The state is a vector that is exposed to an outside agent who can use the state to access the training progress. Ideally one can put as much information as possible into the state design to measure the training progress. However, we found that using a constant (zero) to represent the TPE state can still work as such simple setting allows the trainer to learn a good action quickly. We also test other more informative state representation designs, such as using the last average sampling reward or the normalized sampling count. They can achieve better performance in certain cases. A comparative study of these different designs are provided in Section \ref{stn:experiment}.
 \item \textit{Action}: the action interface comprises three controllable parameters that are exposed to an outside agent who can utilize these actions to control the training progress, as mentioned in Fig. \ref{fig:mbrl_flow}. We represent these parameters as probability values, all defined in the range of $[0, 1]$. Such normalized action range can simplify the design of the trainer agent. Details of the three control parameters will be given subsequently.

\begin{itemize}
 \item Action $a_0$ decides whether one should train the model into a local or global model, which is achieved by controlling the starting state of a new episode when the target controller samples from the real environment. A quality function $\Phi$ is defined to select the starting points, controlled by $a_0$:
 \begin{equation}
 \Phi(s) = a_0 \cdot Q^{\xi}(s,\pi(s)) + (1-a_0) \cdot u_{[0, 1]}, \label{eq_quality_real}
 \end{equation}
 where $Q^{\xi}$ is the value produced by critic network of the target controller, $\pi$ is the current policy, and $u_{[0, 1]}$ is a random number drawn from [0, 1]. With this quality function, we keep sampling random starting points in the physical environment until a high-quality starting point is found, as shown in Algorithm \ref{algo2}. In one way, when $a_0$ approaches to one, initial states with a higher $Q$ value are likely to be selected, which will generate more data with high $Q$ value. When these data are used to train the model, the model will be more accurate in a high $Q$ value subspace, benefiting the local exploitation in it. In the other way, when $a_0$ approaches zero, the quality will be a random number and the starting point will be a random state to favor global exploration.

 \item Action $a_1$ decides whether one should utilize the model in a local or global manner, which is achieved by controlling the starting state of a new episode when the target agent samples from the cyber environment. The starting state of an episode also matters in the cyber environment. For example, we can select a starting state $s$ from the real data buffer $B$. In this case, the subsequent sampling process will be a local search process similar to the imagination process used in \cite{weber2017imagination} and is more likely to generate samples that are of high prediction accuracy as the model has explored nearby samples in the real environment. Alternatively, we can use a data point $s_{rand}$ randomly selected from the state space to favor exploration. It thus can control the trade-off between exploitation and exploration during the sampling process. In our design, $a_1$, with $0 \leq a_1 \leq 1$, represents the probability of choosing starting state $s_0$ from the real data buffer, as
  \begin{equation}
    s_{0}=
\begin{cases}
    s \in B,& \text{if } u_{[0, 1]}\leq a_1\\
    s_{rand},              & \text{otherwise,}
\end{cases}	
\label{eq:a1}
 \end{equation}
where $u_{[0, 1]}$ is a uniformly distributed random number drawn from [0, 1].

\item Action $a_2$ decides how many cyber data are sampled and used in training. For the sampling part, $a_2$ is set to the ratio of the number of real data sampled to the total data sampled (real and cyber) in this training step. Recall that in each step we sample a fixed number $K_r$ of real data samples. $a_2$ controls the number $K_c$ of cyber data to sample in each step by
 \begin{equation}
 K_c = \frac{K_r \cdot (1-a_2)}{a_2}. \label{eq_action1}
 \end{equation}
 The rationale of such design is to bound the action in the range [0, 1], which can ease the design of an agent that interplays with the TPE. The sampled data from the real environment are stored in a real memory buffer, while the cyber data sampled from the cyber environment are stored in the cyber memory buffer.

For the training part, $a_2$ is also used to set the probability of taking a mini-batch from the real data memory buffer in training the target controller. Naturally, $1- a_2$ represents the probability to take a mini-batch from the cyber data buffer. With fixed batch size, if we train with $T_r$ batches of real data, then $T_c$ batches of cyber data are used in this step:
 \begin{equation}
 T_c = \frac{T_r \cdot (1-a_2)}{a_2}. \label{eq_trainingtime}
 \end{equation}
Note that we use only one action to control both the sampling and training process to accommodate some DRL algorithms, such as TRPO, where the sampling and training process cannot be decoupled.
\end{itemize}

 \item \textit{Reward}: The reward interface is used to measure the performance of the target controller. Note that the only reliable information we can get from the training process is the reward data we collected when sampling from the real environment. These reward data can be manipulated into various reward definitions for the trainer; one design of the reward $ r^{\Xi} $ is
\begin{equation}
r^{\Xi} = sign(\bar{r}^{\xi}_{t+1} - \bar{r}^{\xi}_{t}), \label{eq_obj}
\end{equation}
where $\bar{r}^{\xi}_{t+1}$ and $\bar{r}^{\xi}_{t}$ are the respective average sampling reward of the target controller at step $t+1$ and $t$ from real environment. This means, as long as the reward is increasing, the current training action is considered acceptable. Although such a simple design allows the trainer to learn the settings quickly, it may not be effective in all practical cases, especially in the case where the cyber data does not degrade the performance but prolongs the convergence. A more effective order-based reward design is used in the ensemble trainer in Section \ref{stn:ensTrainer}. Note that we can only utilize the reward information received when sampling from the real environment to measure the performance of the target controller. To avoid additional sampling cost, we have to rely on the original sampling process form the real environment, which is why we set the number of real data sampled in each step to be fixed, as otherwise we may not receive a stable evaluation of the target controller.
\end{itemize}

\subsection{Problem Formulation}
Based on the defined TPE environment, the problem to solve in this paper is formulated as follows. Given a target controller to train by a MBRL algorithm with a given maximum number of samples to collect from the physical environment, we encapsulate it into the TPE defined above, and aim to train a trainer in an online manner to maximize the accumulated reward received from this TPE:

\begin{equation}
\max_{\pi^{\Xi}} \sum _{t=1}^{t_{max}} r^{\Xi}(t),  \label{eq_reward}
\end{equation}
where $\pi^{\Xi}$ is the control policy of the trainer, $t_{max}$ is the maximum number of trainer steps when the real data sample budget is consumed. 

\begin{figure}[]
\centering
   \includegraphics[width=0.8\columnwidth] {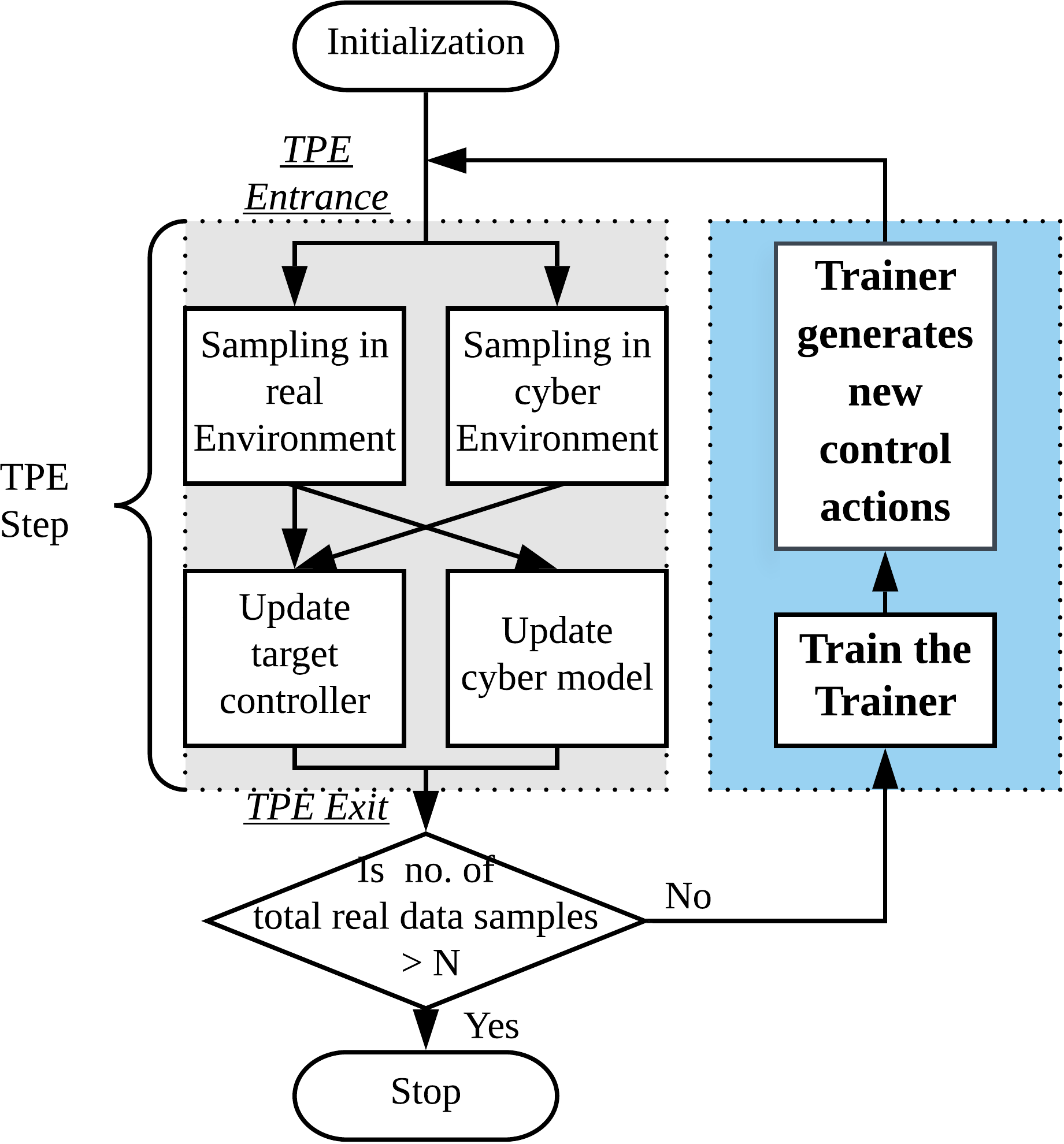}
\caption{Work flow of the uni-head intelligent trainer. In the initialization, we create the corresponding TPE. After that, The training iterates until the total number of real data samples reaching the budget limit $N$.}
\label{fig:uni_flow}
\end{figure}

Note that the problem strictly demands online learning, as re-training from the beginning will incur additional sampling cost. In the following, we will propose different control policy designs and trainer learning methods to accomplish this online learning task.

\section{TTT: Training the Trainer}
In this section we present the outer layer of the RoR architecture, the trainer designs, to tackle the above formulated problem. We first propose the basic intelligent trainer, which utilizes a single DQN controller to do the online learning. Then we propose an enhanced trainer design with multiple trainers to better evaluate the trainer actions, which can even work in some tough situations.

\subsection{Intelligent Trainer}
\label{Trainer}
We design an RL intelligent trainer to optimize control action $a_0$, $a_1$, and $a_2$ of TPE in an online and on-policy manner. The interaction workflow of the trainer with the TPE is shown in Fig. \ref{fig:uni_flow}. At each trainer step  the trainer generates an action setting, with which the TPE advances for one time step. Such a process is equal to the MBRL training process with online parameter adaptation. Note that with such design, only one target controller is involved in training and all trainer actions are tested in a single streamline of training. This ``uni-head'' trainer needs to learn quickly with limited training time steps and samples. Several trainer learning algorithms, like DQN and REINFORCE, can be used to tackle this problem. In the following, we use a DQN controller to demonstrate the trainer design. A comparison of different trainer designs is given in Section \ref{stn:diffdesign}.

We implement a specialized DQN trainer that carries out discretized control actions with a relatively small-scale Q network. At each time step, the trainer evaluates all the actions with the Q network and selects the action with the highest Q value.

The training of the DQN controller follows standard epsilon-greedy exploration \cite{sutton1998reinforcement} strategy. To enhance the training stability, the DQN controller is equipped with a memory, like the replay buffer in DDPG \cite{lillicrap2015continuous}. As such, the trainer can extract good actions from the noisy data received from TPE. During the experiment, we notice that samples from mere one single action could flood the buffer. The homogeneity in actions could prolong or even halt the training of DQN. To solve this problem, for a given action we limit the total number of the samples to $M/|A|$, where $M$ and $|A|$ are the size of buffer and the size of the action set, respectively. If the number of samples for a given action exceeds this limit, a new sample will replace a randomly selected old one.

The pseudo code of the uni-head intelligent trainer is shown in Algorithm \ref{algo2}, with the detailed implementation of the sampling reset procedure in the real/cyber environment shown in Algorithm \ref{algo1}.

\begin{algorithm}[!htb]
\caption{Sampling Reset Procedure}
\label{algo2}
\begin{algorithmic}[1]
\IF {the current sampling environment is the real environment}
    \STATE Initialize data set $D=\emptyset$, quality set $G=\emptyset$.
    \FOR {$i = 1:M_1$}
        \STATE Generate one initial state $s_0$ and compute its quality $\Phi(s_0)$.
        \STATE Append $s_0$ to $D$ and append $\Phi(s_0)$ to $G$.
        \IF {$i>M_2$ and $\Phi(s_0) \ge \max(G)$}
         \STATE Break.
        \ENDIF
    \ENDFOR
    \STATE  Return the last state of $D$.
\ELSE
    \IF {$u_{[0, 1]}<a_1$}
        \STATE Randomly select a state $s$ from the real data memory.
        \STATE Set the cyber environment to state $s$.
        \STATE Return $s$.
    \ELSE
        \STATE Randomly initialize the cyber environment.
        \STATE Return the current state of the cyber environment.
    \ENDIF
\ENDIF

\end{algorithmic}
\end{algorithm}

\begin{algorithm}[!htb]
\caption{Intelligent Trainer Enhanced Model-Based DRL Training Algorithm}
\label{algo1}
\begin{algorithmic}[1]
    \STATE Initialization: initialize the trainer agent (with a DQN network), the training process environment, and the target controller. Initialize real data memory and cyber data memory as an empty set. Sample a small data set of size $o$ to initialize the cyber emulator and initialize the real environment.
    \STATE Set number of total samples generated from real environment $n=0$. Set the maximum number of samples allowed to use as $N$.
    \STATE  $//$\emph{Training Process}:
    \WHILE{$n < N$}
        \STATE Generate action $a$ from the trainer agent.
  \STATE  $//$\emph{One step in TPE}:
   \STATE Train the target controller if there is enough data in its memory buffer.
   \STATE Sample $K_r$ data points from real environment according to the sampling reset Algorithm \ref{algo2}, and append the data to the real data memory.
   \STATE Sample $K_c$ data points from the cyber environment according to the sampling reset Algorithm \ref{algo2}, and append the data to the cyber data memory.
   \STATE Train the dynamic model.
        \STATE Update $n$.
        \STATE Collect the state, action and reward data of TPE.
        \STATE Update the trainer agent.
    \ENDWHILE
\end{algorithmic}
\end{algorithm}

\begin{figure}[]
\centering
   \includegraphics[width=0.95\columnwidth] {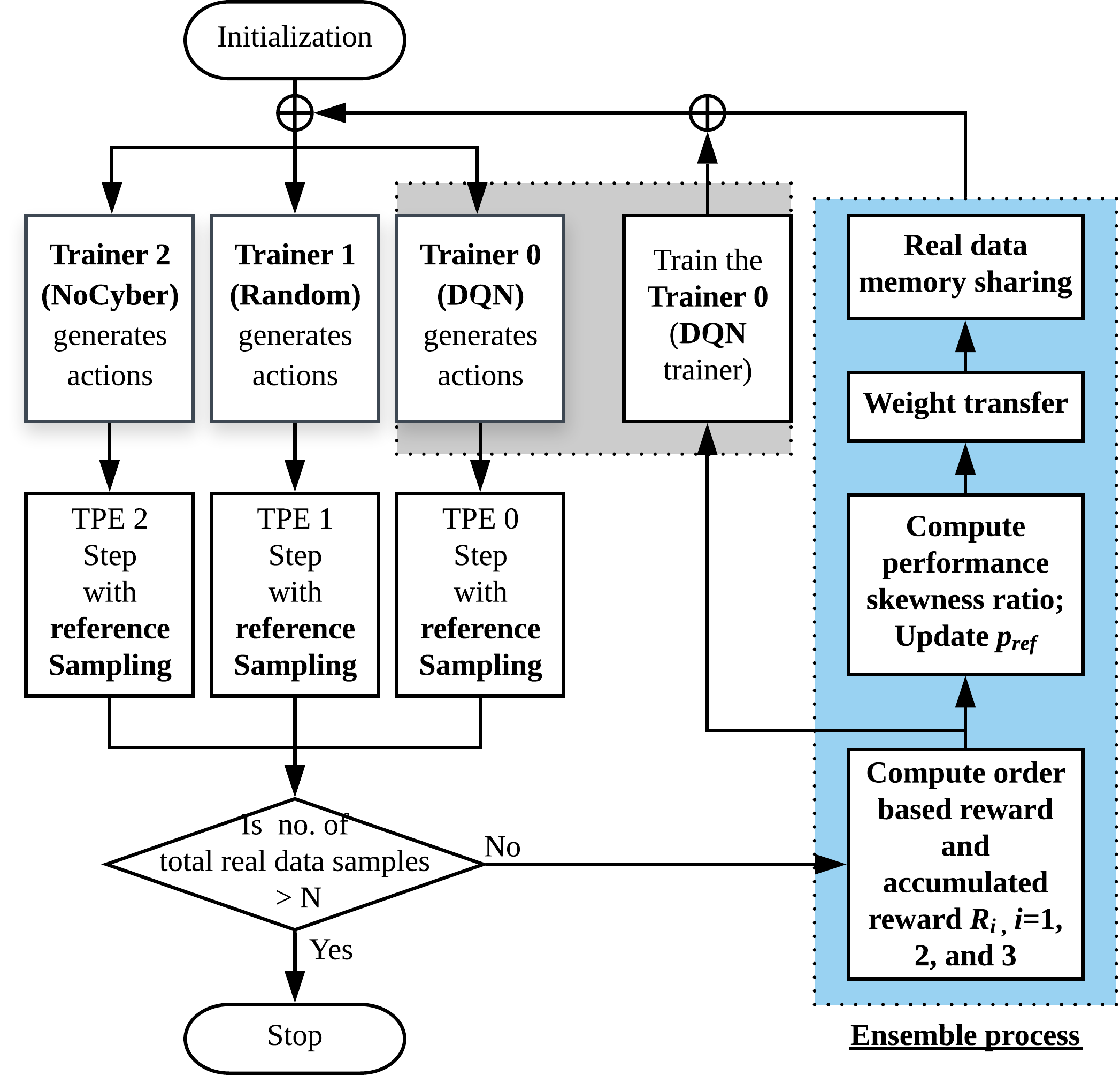}
\caption{Work flow of the ensemble trainer, with the major changes made to the uni-head trainer highlighted in \textbf{bold} font. In the initialization, we create three different trainers and their corresponding TPEs. After that, The training iterates until the total number of real data samples reaching the budget limit $N$. In each training step of ensemble trainer, the original TPE step is revised to include the reference sampling mechanism and after the TPE step, we execute the designed ensemble process including memory sharing, order-based reward calculation and weight transfer.}
\label{fig:ensemble_flow}
\end{figure}

\subsection{Ensemble Trainer}
\label{stn:ensTrainer}
In this subsection, we present a more robust trainer design that learns by comparison to overcome the learning deficiency of uni-head trainer in certain cases. The uni-head trainer, described previously, for some cases cannot adequately assess the quality of the action as all actions are tested in a single streamline. In other words, the actions could be correlated and their quality could become indistinguishable. Also, for actions that generate non-negative reward but could lead to slow convergence or locally optimal policy, the reward function design \eqref{eq_reward} is unable to accurately assess their quality. To address these issues, we propose an ensemble trainer which uses a multi-head training process, similar to the boosted DQN \cite{osband2016deep}. The design rationale is to diversify actions on different trainers without posting additional sampling cost, then evaluate the actions by ranking their performance.

The proposed ensemble trainer consists of three different trainers with the work flow shown in Fig. \ref{fig:ensemble_flow}. For trainer 0, its actions are provided by the intelligent trainer; trainer 1, its actions are provided by a random trainer; trainer 2, it uses only real data, which means setting the three actions to 1, 0, and 0 respectively. The settings in trainer 0 and 1 enable the exploitation and exploration of the action space. Trainer 2 is a normal DRL training process without using the cyber data generated by the dynamic model. The reason we choose to ensemble these three distinct trainers is because they can provide sufficient coverage of different trainer actions and each of them can work well in different cases. Note that it is not a trivial task to have an effective ensemble trainer and at the same time not incurring additional real data cost, as the samples from different trainers can have different quality which may degenerate the ensemble's overall performance. In the following, we propose solutions to solve this issue.

\subsubsection{Real-Data Memory Sharing}
We introduce a memory-sharing scheme to solve the issue of insufficient real data samples for each trainer, as we splitting the whole real data sample budget evenly to all three trainers in the ensemble. The even splitting is necessary for the evaluating of the target controllers of each trainer. It then follows that each trainer only has one-third of the real data samples in training compared with the original uni-head trainer. This will cause significant performance degeneration as will be shown in Section \ref{stn:experiment}. To address this issue, we devise a memory sharing process before the training of the target controller, as shown in Fig. \ref{fig:ensemble_flow}. The memory sharing scheme is a pseudo sampling process executed after each trainer has done its sampling process from real environment and saved these data into their own real memory buffer. Then each trainer will collect the new real data samples from the other trainers. As a result, at each step, each trainer receives $K_r$ new data samples -- the same amount of data as in the uni-head training. Note that with memory sharing, the real data from an underperformed target agent could degrade, even fail the ensemble performance. To solve this problem, we introduce next a reference sampling scheme.

\subsubsection{Reference Sampling}
A reference sampling scheme is proposed to maintain the quality of the real data samples introduced by the memory-sharing mechanism.
The idea behind the reference sampling is to select the best trainer, then to use its target controller for other trainers to sample real data samples with a probability $p_{ref}$. In our algorithm, at the first of every three steps, $p_{ref}$ is forced to set to 0. As such this first step, without reference sampling taking place, serves as an evaluation step for the trainer. In next two steps,
 $p_{ref}$ is determined by the $\min$ function in the following equation.
\begin{equation}
    p_{ref}=
\begin{cases}
    0,& \text{if }  \mod(t^{\Xi},3)==0\\
    \min\{\frac{\phi-\phi_{min}}{\phi_{max}-\phi_{min}} , 1\},  & \text{otherwise}
\end{cases}
\label{eq:pref}
\end{equation}
where $t^{\Xi}$ is the current step number of trainers, and $\phi$ is the skewness ratio, which measures the degree of the outperformance of the best trainer; $\phi_{max}$ and $\phi_{min}$ are the estimated upper and lower bounds respectively. The details of $\phi$ are shown in the weight transfer procedure below. With such design, the better the performance of the best trainer, the higher $p_{ref}$ will be used.

\subsubsection{Order-based Trainer Reward Calculation}
The rewards of the trainers in the ensemble trainer are designed by ordering the performance of different trainers.
After the training process of the target controllers of all trainers, for each trainer we calculate the average sampling reward of its corresponding target controller $\bar{r}^{\xi}_i$ as the raw reward of this trainer. Note that $\bar{r}^{\xi}_i$ is different from the sign reward used in \eqref{eq_reward}.
Next, we sort the tuple ($\bar{r}^{\xi}_0$, $\bar{r}^{\xi}_1$, $\bar{r}^{\xi}_2$) in an ascending order. We then define the index of $c \cdot \bar{r}^{\xi}_i$ in the sorted tuple as the reward $\hat{r}^{\Xi}_i$ of trainer $i$.

The rationale is that if the action of a trainer is good for training, it should help the trainer to achieve better performance (measured by the average sampling reward).

Note that with the above reward design, the trainers will generate three data samples at the trainer level in each step, and all these data will be used to update the intelligent trainer. Due to the reference sampling mechanism, the order information may not correctly measure the performance of the trainers. To solve this issue, we will throw away these samples when $p_{ref}$ is not zero.

\subsubsection{Weight Transfer}
After collection of the trainer reward data, we add a particular weight transfer mechanism to solve the issue that some target agent may fail due to unfavorable trainer actions. The rationale is that after collecting the reward information for a certain large number of steps, we can judge which trainer is currently the best one with high confidence.
In this case, we can transfer the best target agent to the other trainers, such that those trainers who fall behind can restart from a good position. In particular, after the trainer reward data are collected, we examine the number of steps $n_c$ that have been taken since the last weight transfer. If $n_c$ is larger than a threshold $C$, we compute an accumulative reward for each trainer in the last $n_c$ steps as :
\begin{equation}
R_{i}(t^{\Xi}) = \sum_{j \in \{n_c-1, ..., 0\}}\hat{r}^{\Xi}_i(t^{\Xi}-j),
\end{equation}
where $t^{\Xi}$ is the index of current trainer step. The trainer with maximum $R_{i}$ will be set as the best trainer. We then examine if the DQN trainer is the best; if not, we will transfer the weight parameters of the target controller trained by the best trainer to the target controller trained by the DQN trainer.

We also utilize the accumulated trainer reward to detect whether the best trainer is significantly better than other trainers. We calculate a performance skewness ratio to measure the degree of the outperformance of the best trainer:
\begin{equation}
\phi = \frac{R_{b}-R_{m}}{R_{b}-R_{w}},
\end{equation}
where $R_{b}$, $R_{m}$ and $R_{w}$ are the best, median and worst $R_i$ of the three trainers, respectively. The skewness ratio is used to determine the $p_{ref}$ as shown above.

Algorithm \ref{algo3} shows the operational flow of the ensemble trainer. In summary, the ensemble trainer evaluates the quality of the actions by sorting the rewards received by target controllers.  It can maintain the training quality by memory sharing scheme, without incurring additional sampling cost. It can maintain the sample quality by reference sampling. It can recover an underperformed trainer from poor actions. Though saving on the sampling cost, the ensemble trainer requires three times the training time. The increased training time can be partially reduced by the early stop of some underperformed trainers when necessary.

\begin{algorithm}[!htb]
\caption{Ensemble Trainer Algorithm}
\label{algo3}
\begin{algorithmic}[1]
    \STATE Initialization: initialize the three trainer agents and the corresponding training process environments, along with the target controllers. Run the initialization process for each trainer. Initialize the best player to be NoDyna trainer and the probability to use best player to sample is $p_{ref}$.
    \STATE Set number of total samples generated from real environment $n=0$. Set maximum number of samples allowed to use as $N$.
    \STATE  $//$\emph{Training Process}:
    \WHILE{$n < N$}
     \FOR {trainer $i$ $\in$ {0, 1, 2}}
         \STATE Generate action $a$ from the trainer agent.
   \STATE  $//$\emph{One step in TPE}:
    \STATE Execute memory sharing procedure.
    \STATE Train the target controller if there is enough data in its memory buffer.
    \STATE Sample $K_r/3$ data points from real environment with reference sampling probability $p_{ref}$, and append the data to the real data memory.
    \STATE Sample data from cyber environment according to the trainer action, and append the data to the cyber data memory.
    \STATE Share the real data memory across all trainers.
    \STATE Train the dynamic model of the current trainer.
         \STATE Update $n$.
         \STATE Collect the state, action and raw reward data of TPE.
        \ENDFOR
        \STATE Compute reward for each trainer from the raw reward data and calculate the accumulative reward $R_i$ for trainers $i = 0, 1, 2$.
        \STATE Store TPE data of all three trainers into the DQN memory to train the intelligent trainer.
        \STATE Update the trainer agents.
        \STATE Execute Algorithm \ref{algo4} to do performance skewness analysis and weight transfer, update $p_{ref}$.
    \ENDWHILE
\end{algorithmic}
\end{algorithm}

\begin{algorithm}[!htb]
\caption{Performance Skewness Analysis Procedure}
\label{algo4}
\begin{algorithmic}[1]

\IF {$n_c > C$}
    \STATE Compute accumulative reward of trainer $i$ as $R_i$ for $i = 0, 1, 2$.
    \STATE Update best trainer index as $\argmax_i(R_i)$.
    \STATE Compute the skewness ratio $\phi$ for the best player.
    \STATE Update best player reference probability $p_{ref}$ according to \eqref{eq:pref}.
    \IF {DQN trainer is not the best trainer}
        \STATE Do weight transfer from the best trainer to DQN trainer.
    \ENDIF
    \STATE Reset $n_c=0$.
\ENDIF

\end{algorithmic}
\end{algorithm}

\section{Numerical Evaluations}
\label{stn:experiment}
In this section, we evaluate the proposed intelligent trainer and ensemble trainer for five different tasks (or cases) of OpenAI gym: Pendulum (V0), Mountain Car (Continuous V0), Reacher (V1), Half Cheetah (\cite{HW4}), and Swimmer (V1).

\subsection{Experiment Configuration}
For the five test cases, different target controllers with promising published results are used: DDPG for Pendulum and Mountain Cars; TRPO for Reacher, Half Cheetah, and Swimmer. The well-tuned parameters of open-sourced codes \cite{TRPO_Example_Git}\cite{baselines} are used for the hyper-parameters settings of the target controller (including $K_r$ and $T_r$, as defined in Section \ref{stn:approach}. Simple neural networks with guideline provided in [25] are used for the cyber models. As our experiments have shown, it is very useful to normalize both input and output for the dynamic model. In this paper, we use the normalization method provided by \cite{TRPO_Example_Git}, in which the mean and standard deviation of the data is updated during the training process. For hyperparamters $M_1$ and $M_2$ used in the reset procedure in Algorithm \ref{algo2}, we set $M_1=50$ and $M_2=5$ respectively, which indicates that we have maximum and minimum trial numbers 50 and 5 respectively.

\begin{table*}[]
\centering
\caption{Configurations of different algorithms.}
\label{tbl:trainer_conf}
\resizebox{\textwidth}{!}{%
\begin{tabular}{c|c|c|c|c|c|c|c|c|c}
\hline
 & \multicolumn{3}{c|}{Baseline algorithms} & \multicolumn{6}{c}{Intelligent trainers} \\ \hline
 & NoCyber & Fixed & Random & DQN & DQN-5 actions & DQN-larger memory & REINFORCE & DQN-TPE V1 & DQN-TPE V2 \\ \hline
Trainer type & None & None & None & DQN & DQN & DQN & \textbf{REINFORCE} & DQN & DQN \\ \hline
Action & (1, 0, 0) & (0.6, 0.6, 0.6) & $a_i \in \{0.2, 1.0\}$ & $a_i \in \{0.2, 1.0\}$ & \boldmath$a_i \in \{0.2,  0.4 , 0.6 , 0.8 , 1.0\}$ & $a_i \in \{0.2, 1.0\}$ & $a_i \in \{0.2, 1.0\}$ & $a_i \in \{0.2, 1.0\}$ & $a_i \in \{0.2, 1.0\}$ \\ \hline
Data source & Real & Real \&  Cyber & Real \&  Cyber & Real \&  Cyber & Real \&  Cyber & Real \&  Cyber & Real \&  Cyber & Real \&  Cyber & Real \&  Cyber \\ \hline
Memory size & - & - & - & 32 & 32 & \textbf{2000} & - & 32 & 32 \\ \hline
TPE state & - & - & - & Constant & Constant & Constant & Constant & \textbf{Last sampling reward} & \textbf{Real sample count} \\ \hline
\end{tabular}%
}
\end{table*}

\begin{table}[]
\centering
\caption{Number of total TPE steps for different tasks.}
\label{tbl:TPE steps}
\begin{tabularx}{1.0\columnwidth}{X|X|X|X|X|X}
\hline
              & Pendulum & Mountain Car & Reacher & Half Cheetah & Swimmer \\ \hline
TPE Steps     & 1000     & 30000                  & 1000    & 400          & 200     \\ \hline
\end{tabularx}%

\end{table}

\subsection{Comparison of Uni-Head Intelligent Trainer with Baseline Algorithms}
Multiple variants of the uni-head intelligent trainer are compared with baseline algorithms. There are three baseline algorithms and four intelligent trainers. Their designs are summarized in Table \ref{tbl:trainer_conf}. The three baseline algorithms are:
\begin{itemize}
\item The NoCyber trainer is a standard DRL training process without using cyber data.
\item The Fixed trainer follows the standard MBRL, with all actions set to 0.6 throughout the training process.
\item The Random trainer outputs action 0.2 or 1.0 with equal probability. The same action values will be used by the DQN trainer. These values are picked such that an extensive amount of cyber data can be used in the training, for example, when $a_2$ is set to 0.2, the amount of cyber data sampled is five-time the real data sampled. The value 0.2 is chosen without any tuning, i.e., it is not tuned to make DQN trainer work better. Our focus is not to find out the best settings of these parameters (as it will vary in practice), but to figure out if the proposed trainer can select the better action among the predefined action value set.
\end{itemize}

We notice that, for some tasks, the total number of steps of the TPE is only 200, as shown in Table. \ref{tbl:TPE steps}. To simplify the learning process, we discretize each dimension of the trainer action.

The four intelligent trainers are:
\begin{itemize}
\item DQN trainer. The trainer action chooses from two values of 0.2 and 1.0 like the Random trainer. That is, $a_i\in \{0.2, 1\}$ for $i=0,1,2$. The DQN controller is trained with a memory buffer of size 32. At each time steps, four randomly selected batches of batch size eight are used to update the controller. For exploration purpose, the epsilon-greedy method is used, with the first 10\% of the trainer steps for epsilon-greedy exploration by setting final epsilon to 0.1. Note that the setting 0.6 used in Fixed trainer is the expected mean of the actions from intelligent trainer if the trainer predicts uniformly random actions.
\item DQN-5 actions. To test the effect of more action values in the action discretization, we introduce a second trainer, by selecting five values from $\{0.2, 0.4, 0.6, 0.8, 1\}$.
\item DQN-larger memory. To test the impact of larger trainer memory, we introduce a third intelligent trainer with memory size of 2000. In this case more trainer samples are stored and relatively older historical data are used in the training the DQN controller.
\item REINFORCE. The fourth intelligent trainer is the same to DQN trainer except the DQN controller is replaced by a REINFORCE controller. REINFORCE algorithm requires data of multiple episodes to train, we manually set five steps (manually tuned) of TPE as an episode.
\end{itemize}

The configurations for these algorithms are summarized in Table \ref{tbl:trainer_conf}.

\begin{figure*}[]
\centering
\subfigure[Pendulum V0]{
   \includegraphics[width=0.3\textwidth] {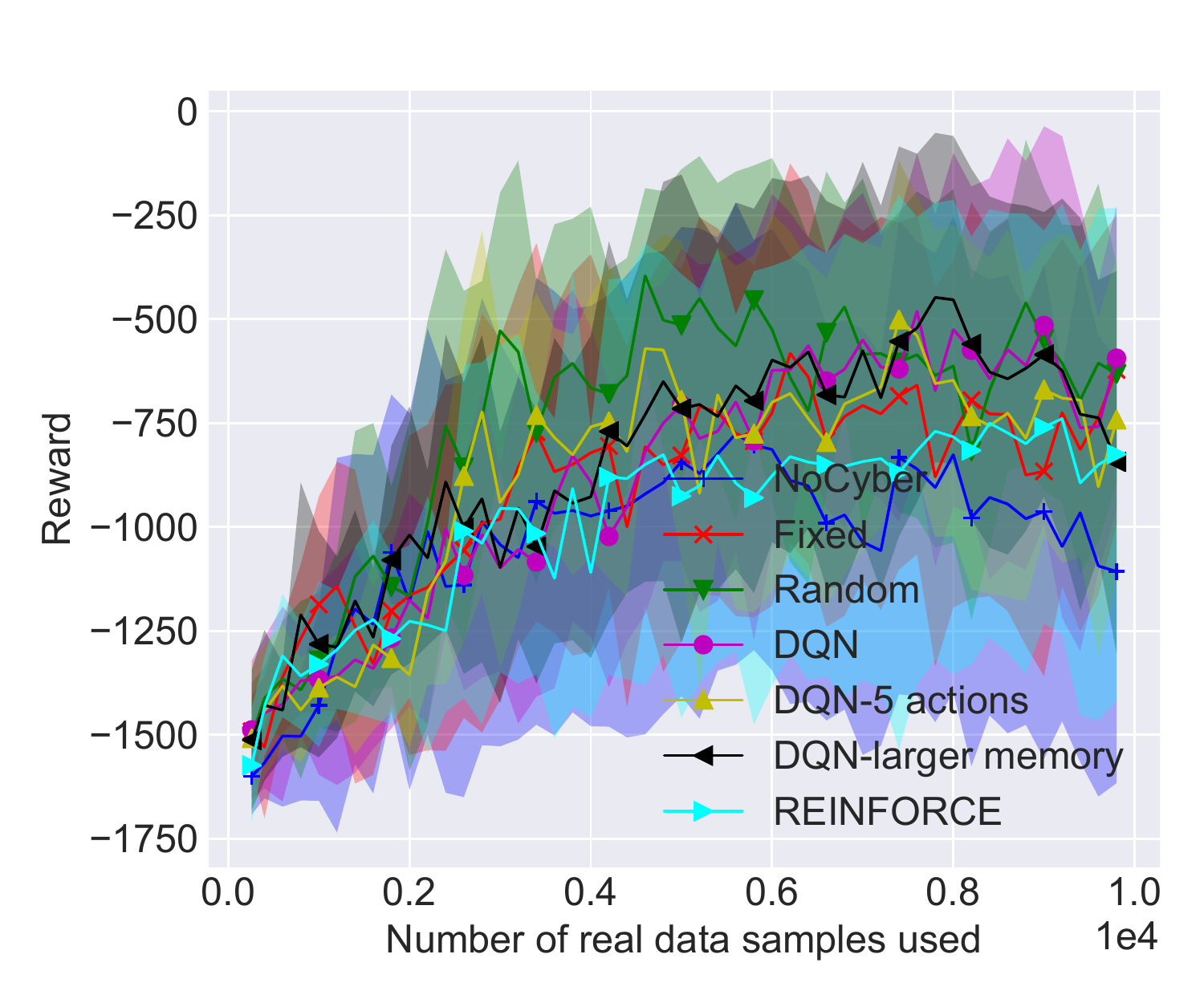}
}
\subfigure[Mountain Car Continuous V0]{
   \includegraphics[width=0.3\textwidth] {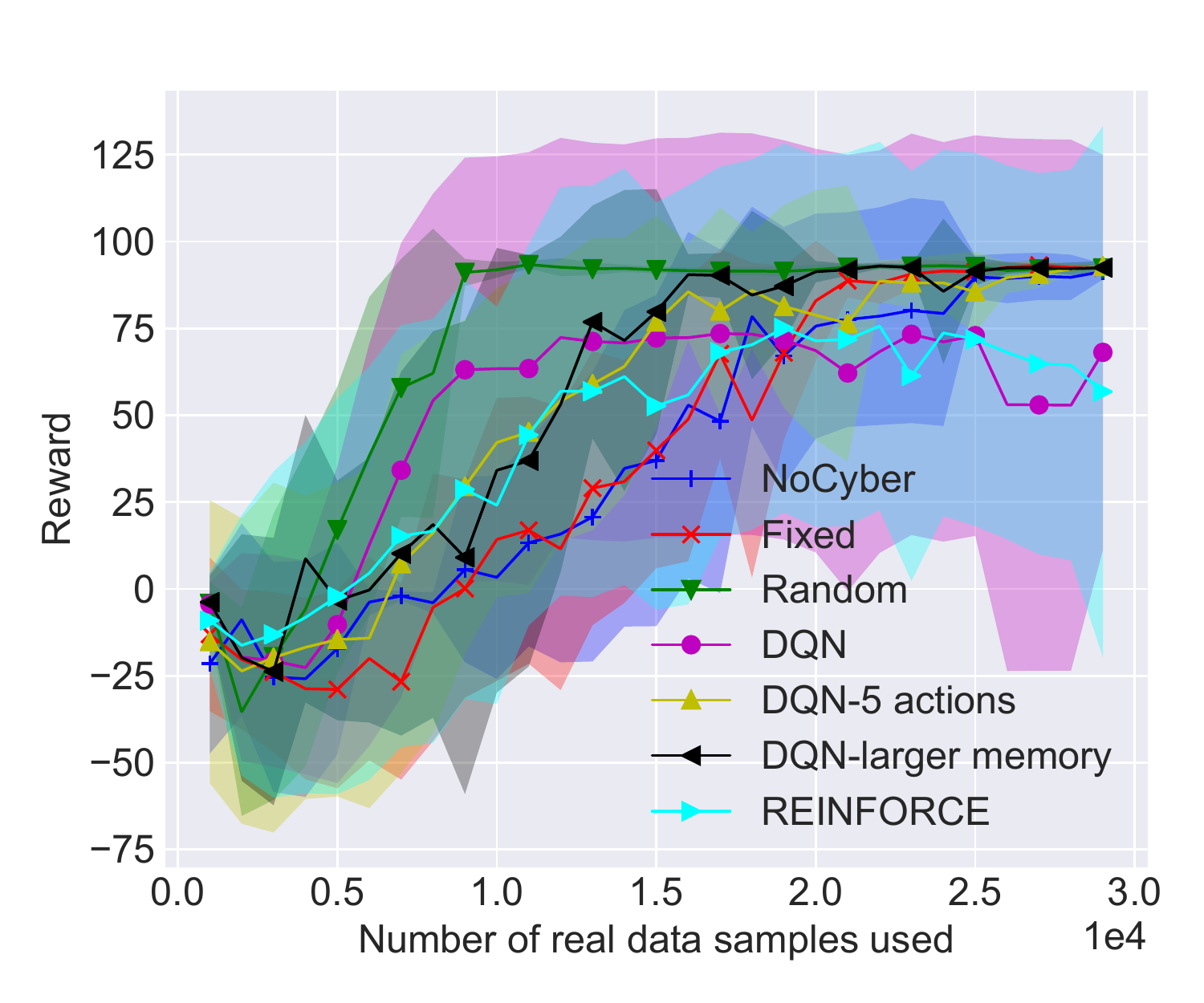}
}
\subfigure[Reacher V1]{
   \includegraphics[width=0.3\textwidth] {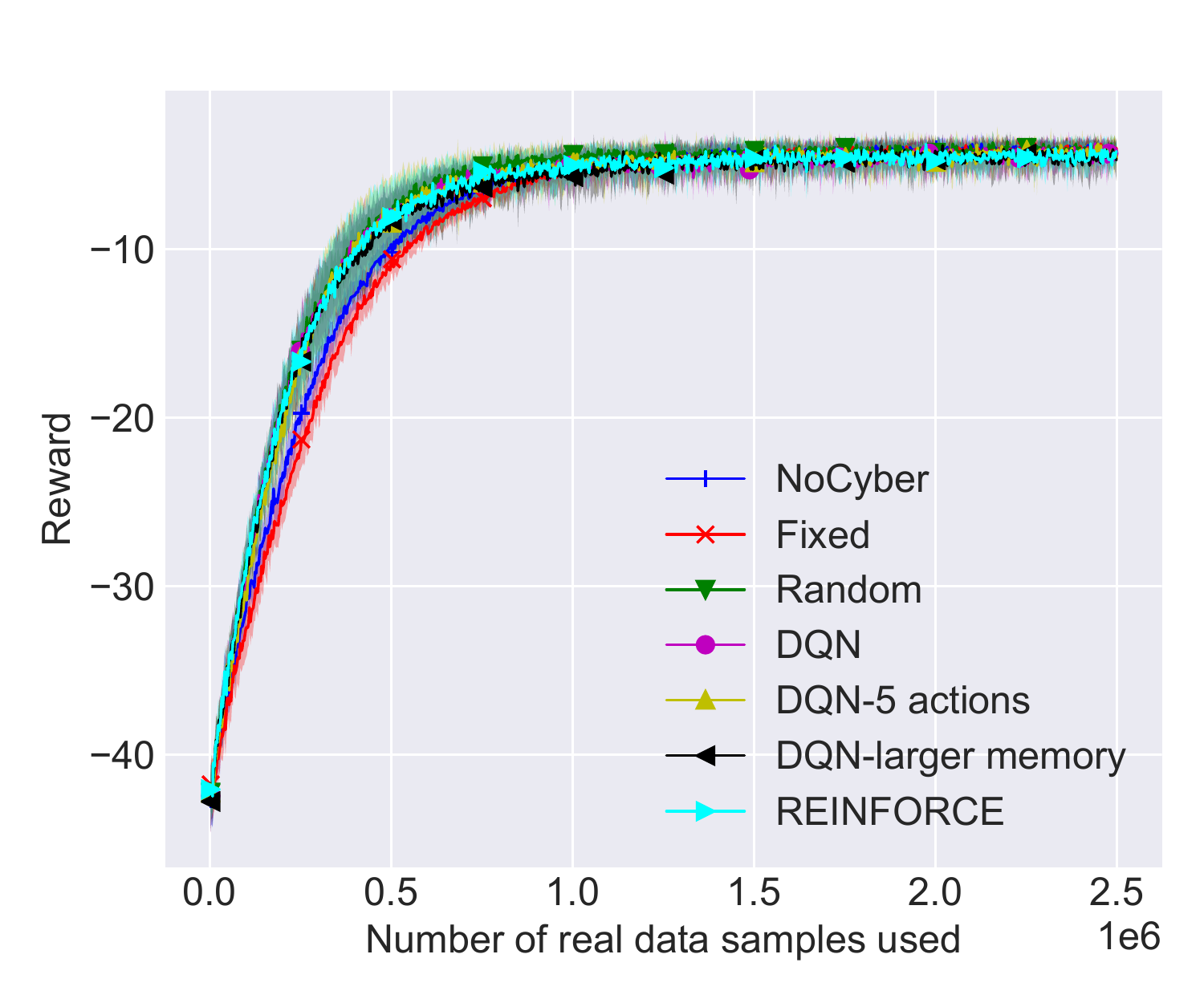}
}

\subfigure[Half Cheetah]{
   \includegraphics[width=0.3\textwidth] {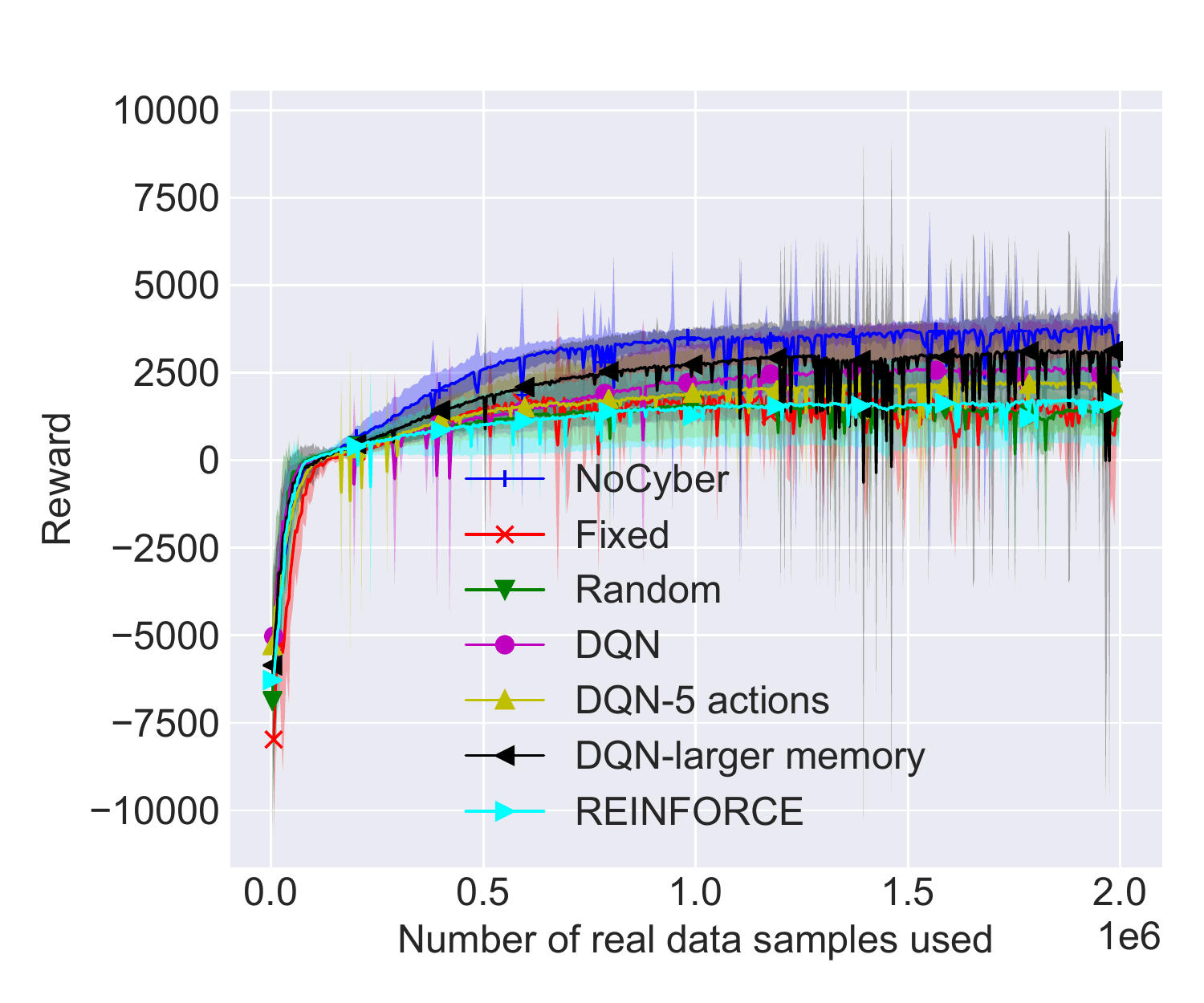}
}
\subfigure[Swimmer V1]{
   \includegraphics[width=0.3\textwidth] {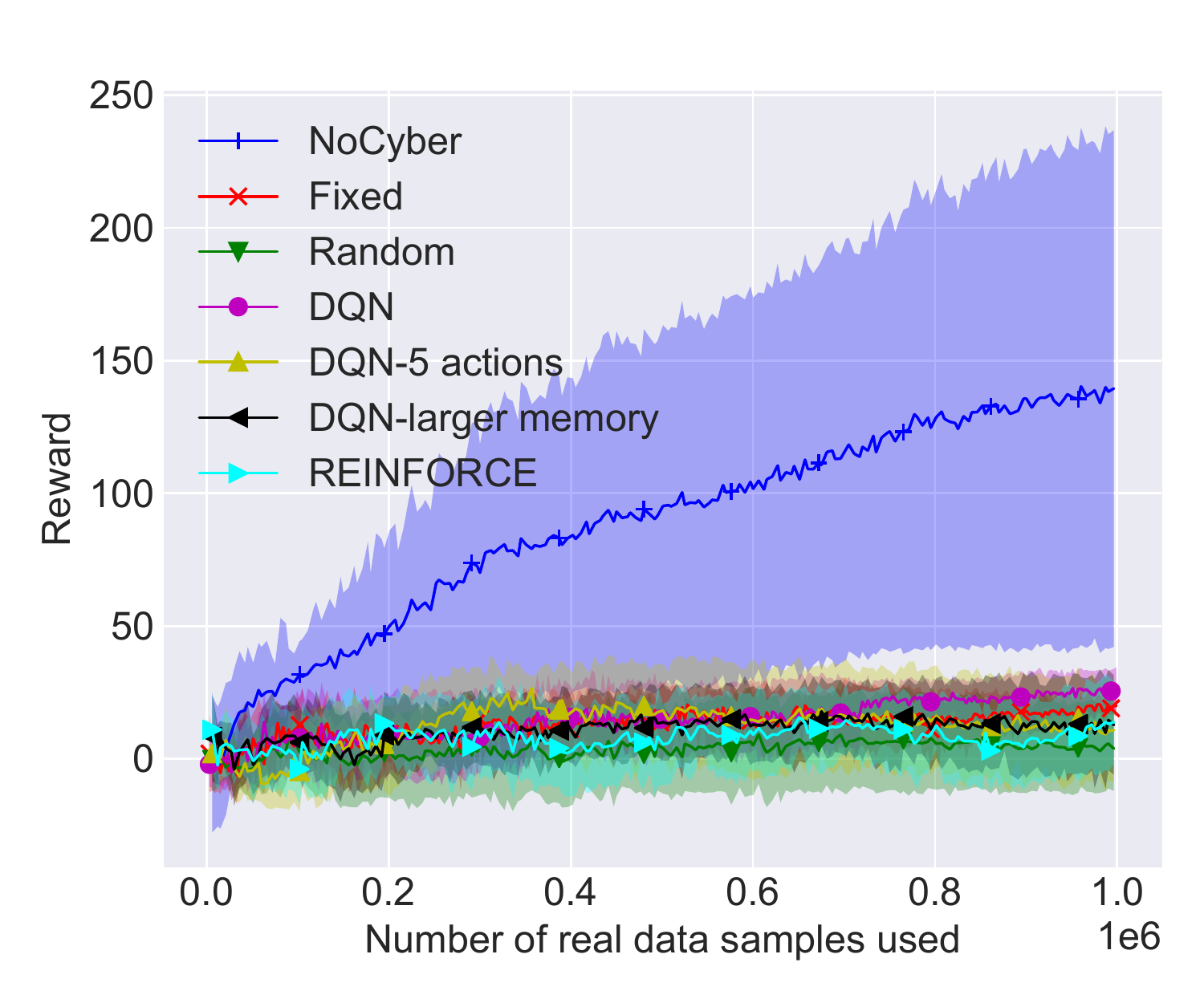}
}
\subfigure[Trainer action $a_2$]{
   \includegraphics[width=0.3\textwidth] {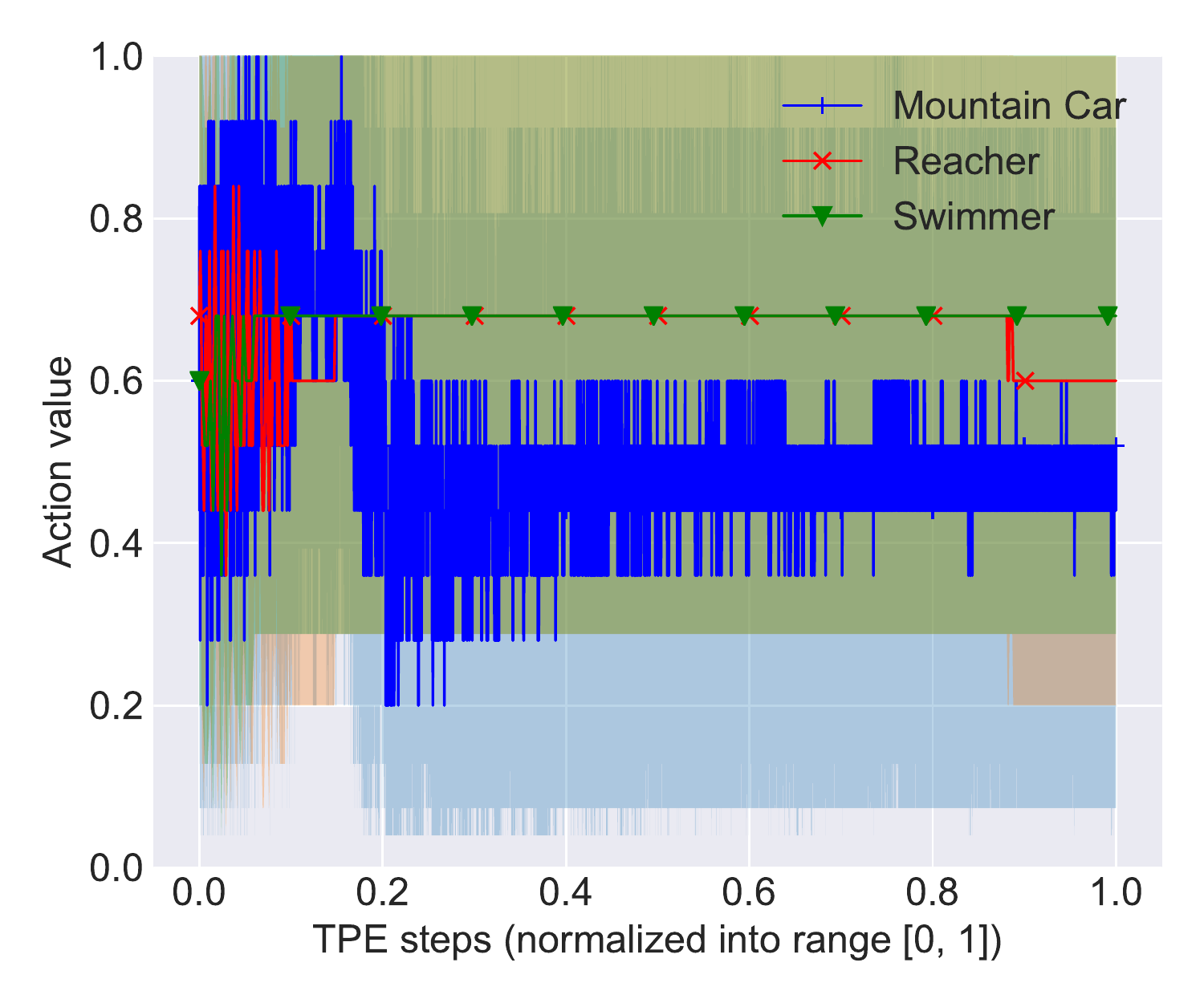}
}
\caption{Accumulative rewards of different \textbf{uni-head trainer} designs for different tasks in (a)-(e). The curves show the average accumulative reward while the shaded region shows the standard deviation of the reward in ten independent runs. The proposed uni-head trainer shows its adaptability (better than the Fixed trainer) but may fail in certain cases like Swimmer. (f) shows the mean action $a_2$ taken by DQN trainer on tasks of Mountain Car, Reacher, and Swimmer. }
\label{fig:results_5case}
\end{figure*}

The test results of three baseline trainers and four intelligent trainers are shown in Fig. \ref{fig:results_5case}. We obtain the test results by periodically evaluating the target controller in an isolated test environment. This ensures that data collection from the test environment will not interfere with the training process. In other words, none of the data collected from the test environment is used in the training. We observe that:

\begin{itemize}
 \item The tasks of Pendulum, Mountain Car, and Reacher can benefit from cyber data used in training. For tasks of Half Cheetah and Swimmer, NoCyber trainer performs significantly better than trainers using cyber data. This indicates that using the cyber data may not be always beneficial. Thus, the use of cyber model should be considered carefully.
  \item In most tasks, the intelligent trainer performs better than the Fixed trainer. For example, DQN-5 actions performs better than Fixed trainer for the tasks of Mountain Car, Reacher, and Half Cheetah, and performs similarly for the tasks of Pendulum and Swimmer. This indicates the viability of the intelligent trainer.

 \item For the tasks of Pendulum and Mountain Car, the Random trainer performs the best. This can be attributed to the fact that adding more noises would encourage exploration. For example, to achieve better performance, the Mountain Car requires more exploration to avoid local optimum that could lead the target agent to unfavorable searching directions. We also observe that the performance of DQN-5 actions is more stable than that of DQN, due to the increased dimension of action space that improves the training diversity. We argue that even the DQN trainer is no better than the Random trainer in these tasks, the DQN trainer is still learning something. The reason is that we are trying to learn a fixed good action through DQN trainer, which means that the DQN trainer will not be able to provide the randomness which proves to be good in these tasks. Also we can observer that for the Half Cheetah task, the DQN trainer is much better than the Random trainer. This suggests that the DQN trainer can indeed learn in an online manner.

 \item We further examine the effect of using cyber data when it seems not working. For the Half Cheetah, we examine the results of multiple independent runs and cyber data causes instability in performance, resulting in higher variance and low mean reward in ten independent tests. For Swimmer, the poor performance with cyber data is due to a special feature that the first two dimensions are linearly correlated in its state definition. The trained cyber model in this case is unable to correctly identify this feature and predict the state transition. Our results show that even incorporating 10\% cyber data in training, severe performance degradation can occur. When cyber data are used, the target controller can be trapped by a local optimum that is difficult to recover from. We resolve this issue by using ensemble trainer.
\end{itemize}

\begin{table}[]
\caption{Accumulative rewards of different trainer variants when using different trainer and TPE designs.}
\label{tbl:variants_test}
\resizebox{\columnwidth}{!}{%
\begin{tabular}{l|c|c|c|c|c}
\hline
\multicolumn{1}{c|}{Variants} & \multicolumn{1}{l|}{Pendulum} & \multicolumn{1}{l|}{Mountain Car} & \multicolumn{1}{l|}{Reacher} & \multicolumn{1}{l|}{Half Cheetah} & \multicolumn{1}{l}{Swimmer} \\ \hline
DQN & -43323 & 1434.59 & -7846 & 696492 & \textbf{4918} \\ \hline
DQN-5 actions & -43204 & 1493.88 & -7724 & 985847 & \textbf{2473} \\ \hline
DQN-larger memory & \textbf{-41329} & 1615.98 & -7831 & \textbf{1354488} & 2142 \\ \hline
DQN-TPE V1 & \textbf{-41869} & \textbf{1849.96} & \textbf{-7456} & 868597 & 1522 \\ \hline
DQN-TPE V2 & -46533 & \textbf{1826.19} & \textbf{-7478} & \textbf{1172288} & 2233 \\ \hline
\end{tabular}%
}
\end{table}

To analyze the behavior of the trainer, we show in Fig. \ref{fig:results_5case}(f) the actions taken by the DQN trainer for the tasks of Mountain Car, Reacher, and Swimmer during the training process. We observe that for Mountain Car, the mean value of $a_{0}$ fluctuates around 0.5. This agrees with our observation that for the Mountain Car, random baseline algorithm performs the best. For Reacher and Swimmer, the trainer quickly learns to use more of the real data, with the mean value of action $a_2$ eventually reaching to larger than 0.6. This again indicates the viability of the trainer. Note that for Swimmer, even the mean value of action $a_2$ is larger than 0.6, the performance of the target controller is still very poor (Fig. \ref{fig:results_5case}) due to training process' sensitivity to cyber data. This again verifies the necessity of an ensemble trainer that can quickly recover from degraded performance during training.

\begin{figure*}[]
\centering
\subfigure[Pendulum V0]{
   \includegraphics[width=0.3\textwidth] {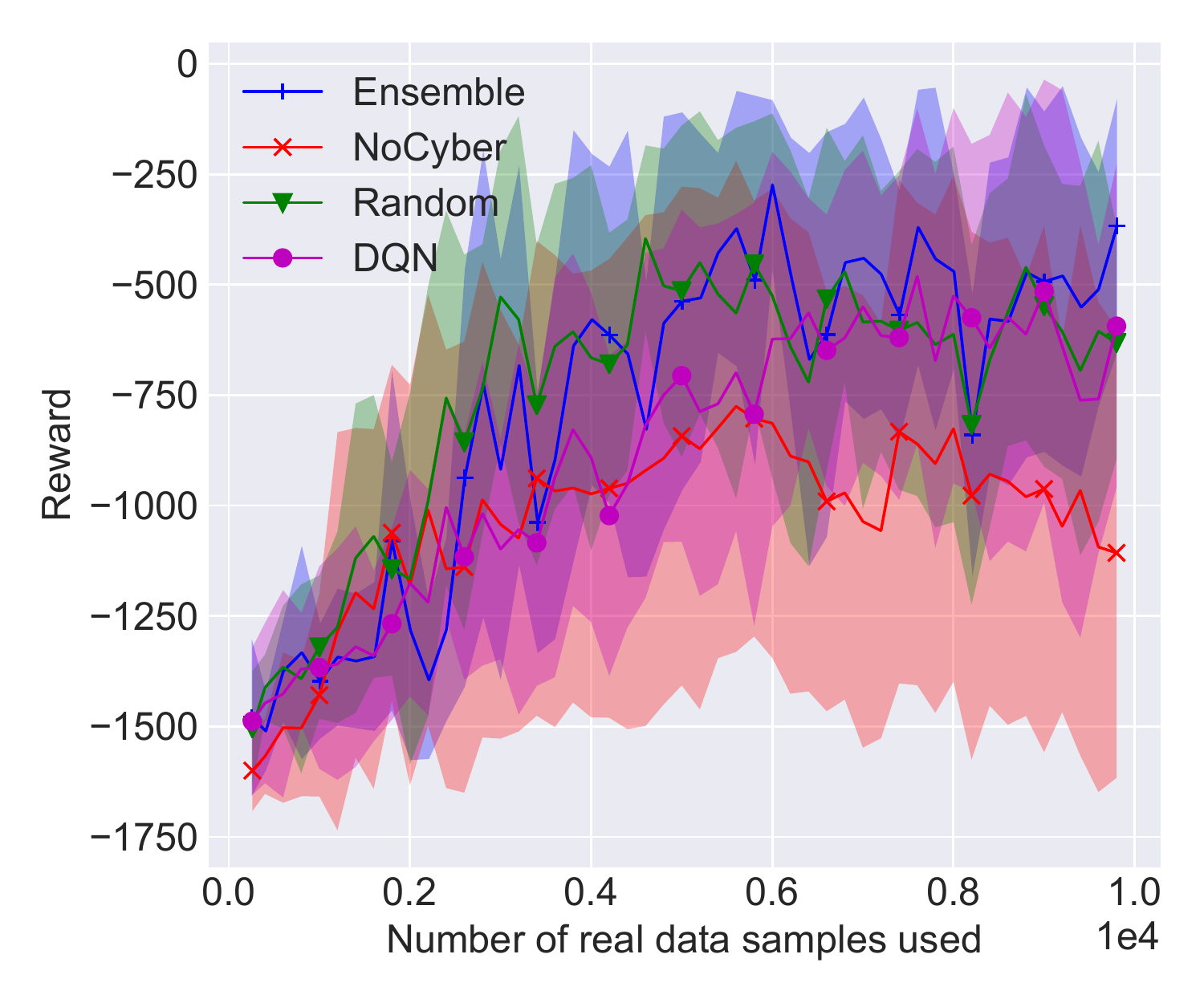}
}
\subfigure[Mountain Car Continuous V0]{
   \includegraphics[width=0.3\textwidth] {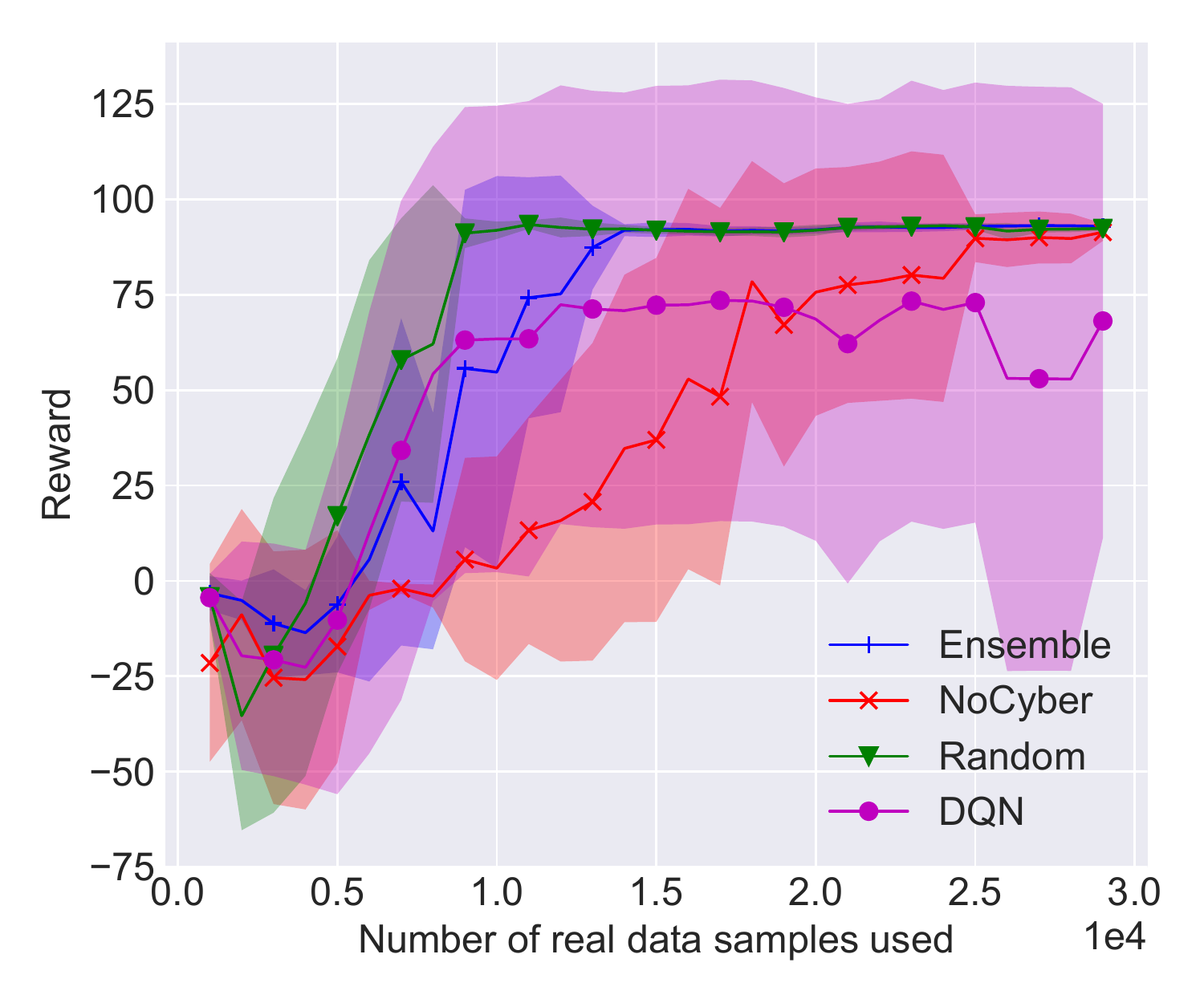}
}
\subfigure[Reacher V1]{
   \includegraphics[width=0.3\textwidth] {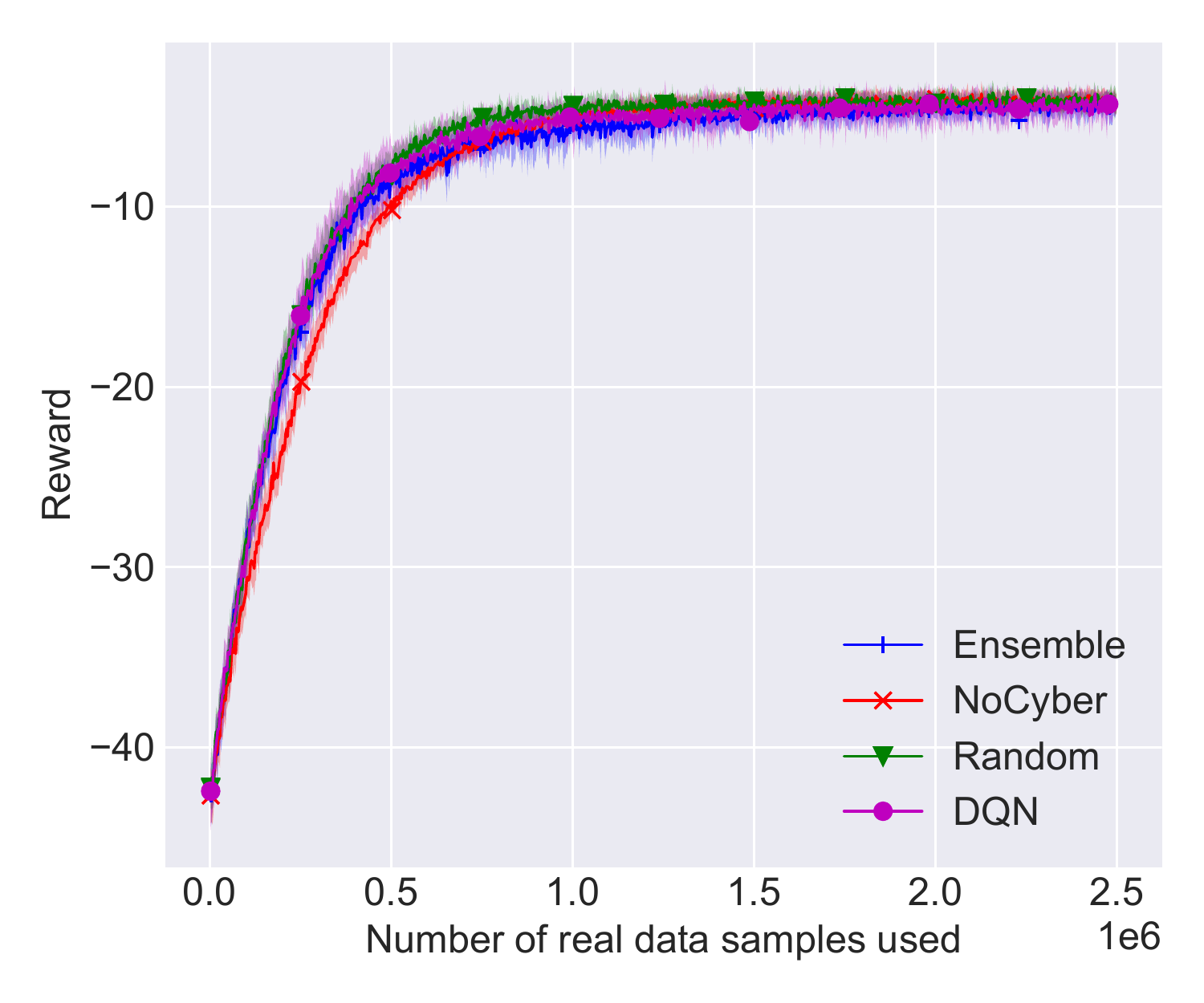}
}

\subfigure[Half Cheetah]{
   \includegraphics[width=0.3\textwidth] {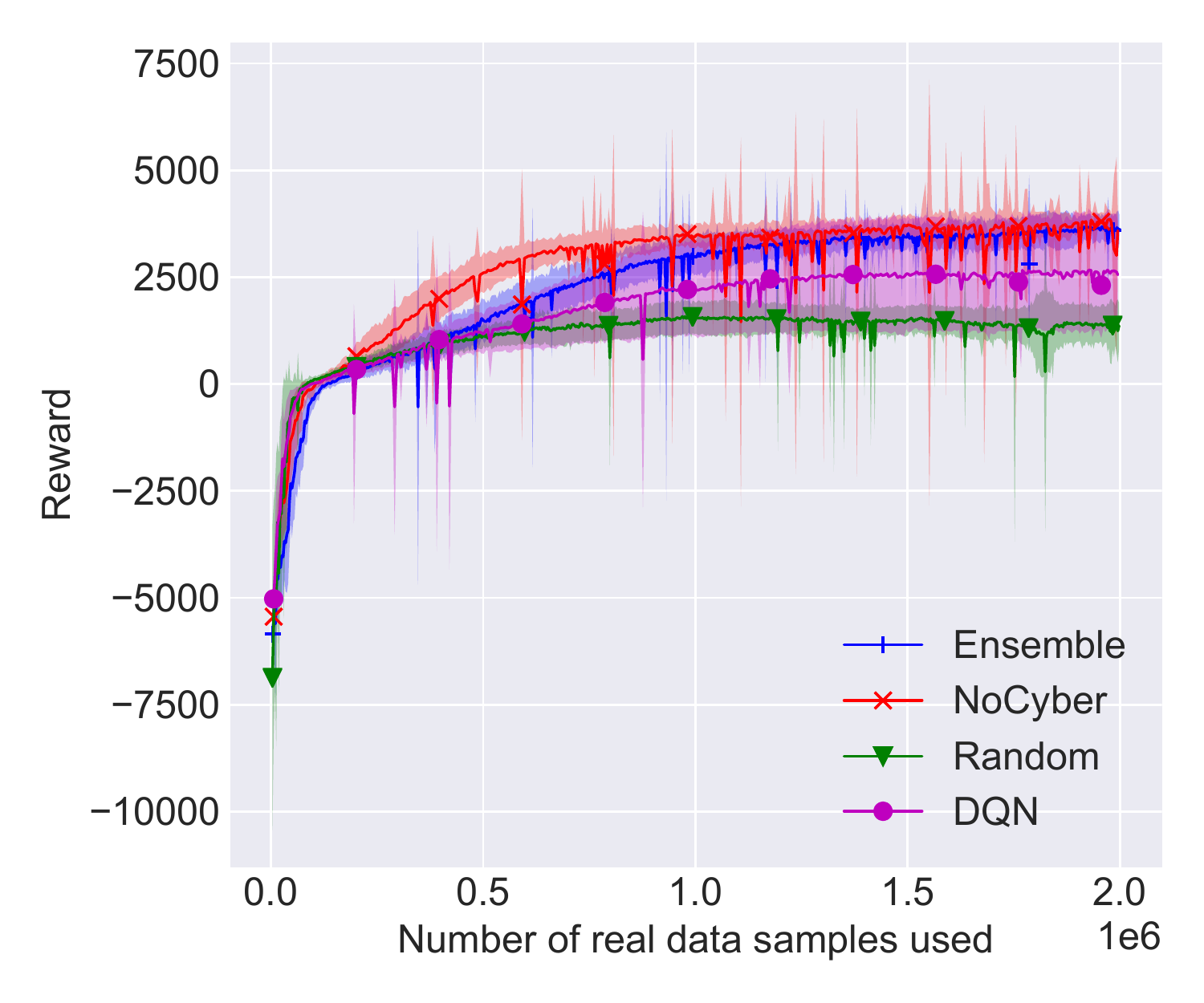}
}
\subfigure[Swimmer V1]{
   \includegraphics[width=0.3\textwidth] {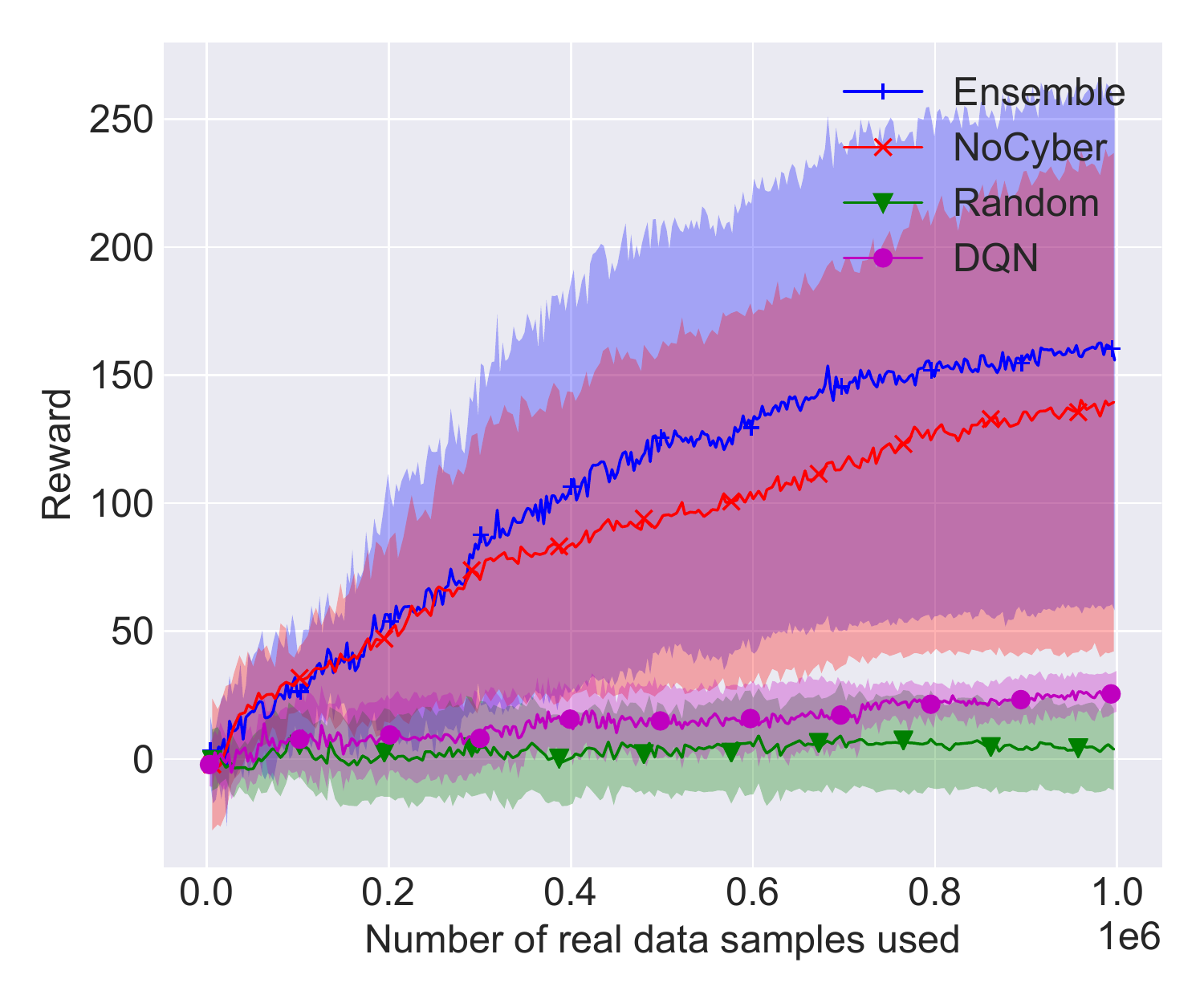}
}
\subfigure[Trainer action $a_2$]{
   \includegraphics[width=0.3\textwidth] {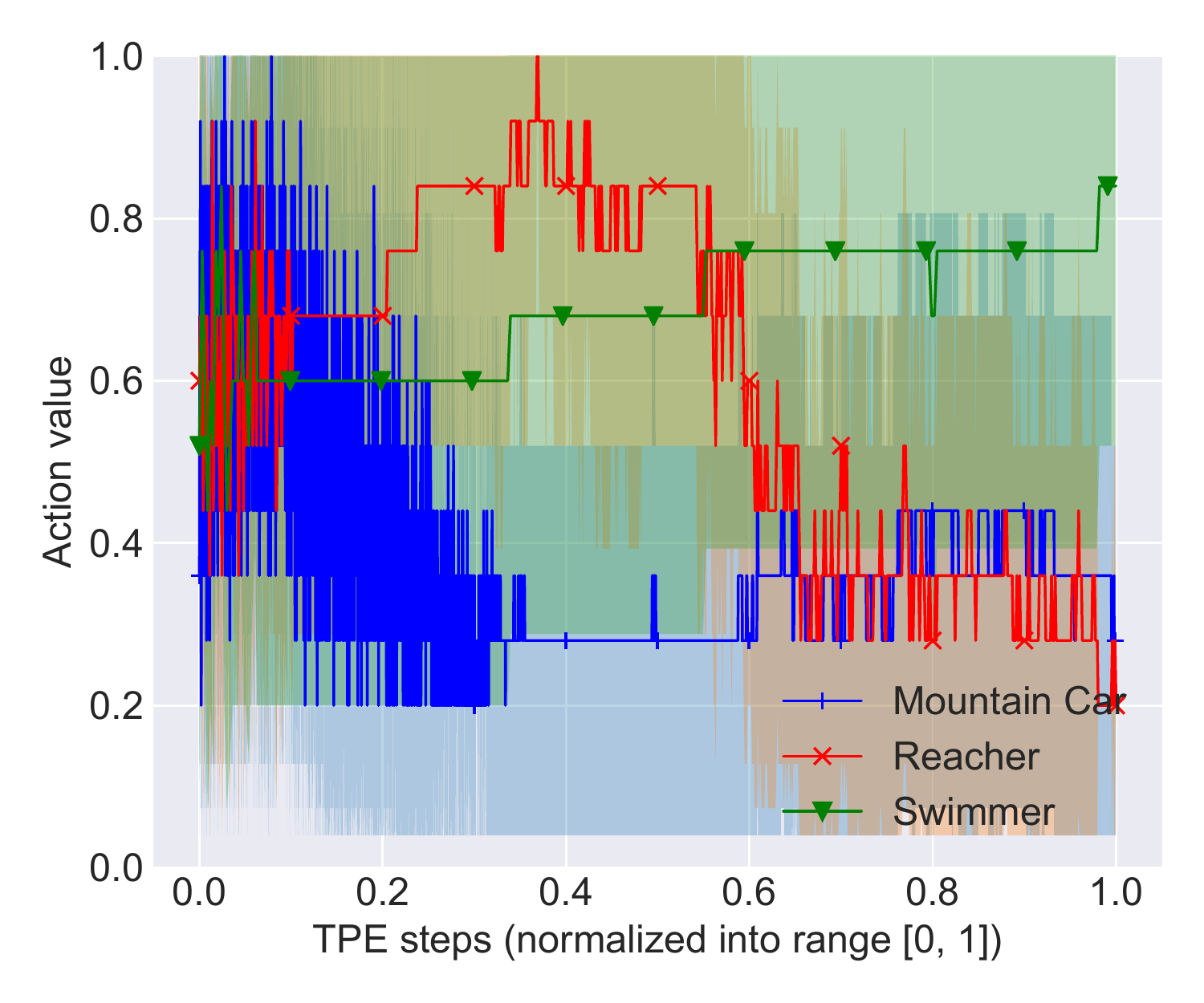}
}
\caption{Accumulative rewards of \textbf{ensemble trainer} for different tasks in (a)-(e). The proposed ensemble design shows close-to-optimal or even better performance on all cases. (f) shows the mean action $a_2$ taken by the DQN trainer in the ensemble trainer for Mountain Car, Reacher, and Swimmer.}
\label{fig:ensemble_results}
\end{figure*}

\begin{figure*}[!t]
\centering
\subfigure[Mountain Car Continuous V0]{
   \includegraphics[width=0.3\textwidth] {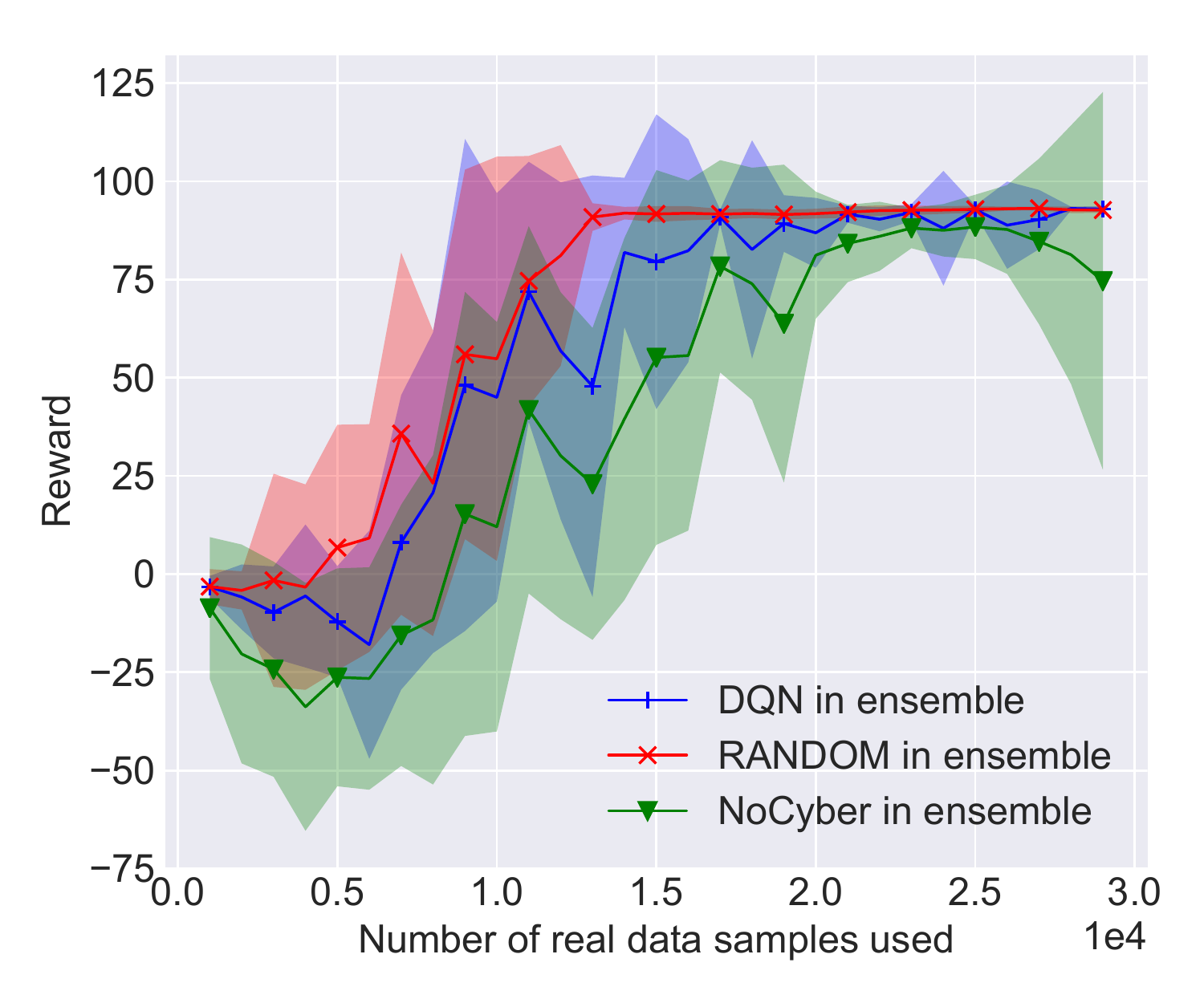}
}
\subfigure[Reacher V1]{
   \includegraphics[width=0.3\textwidth] {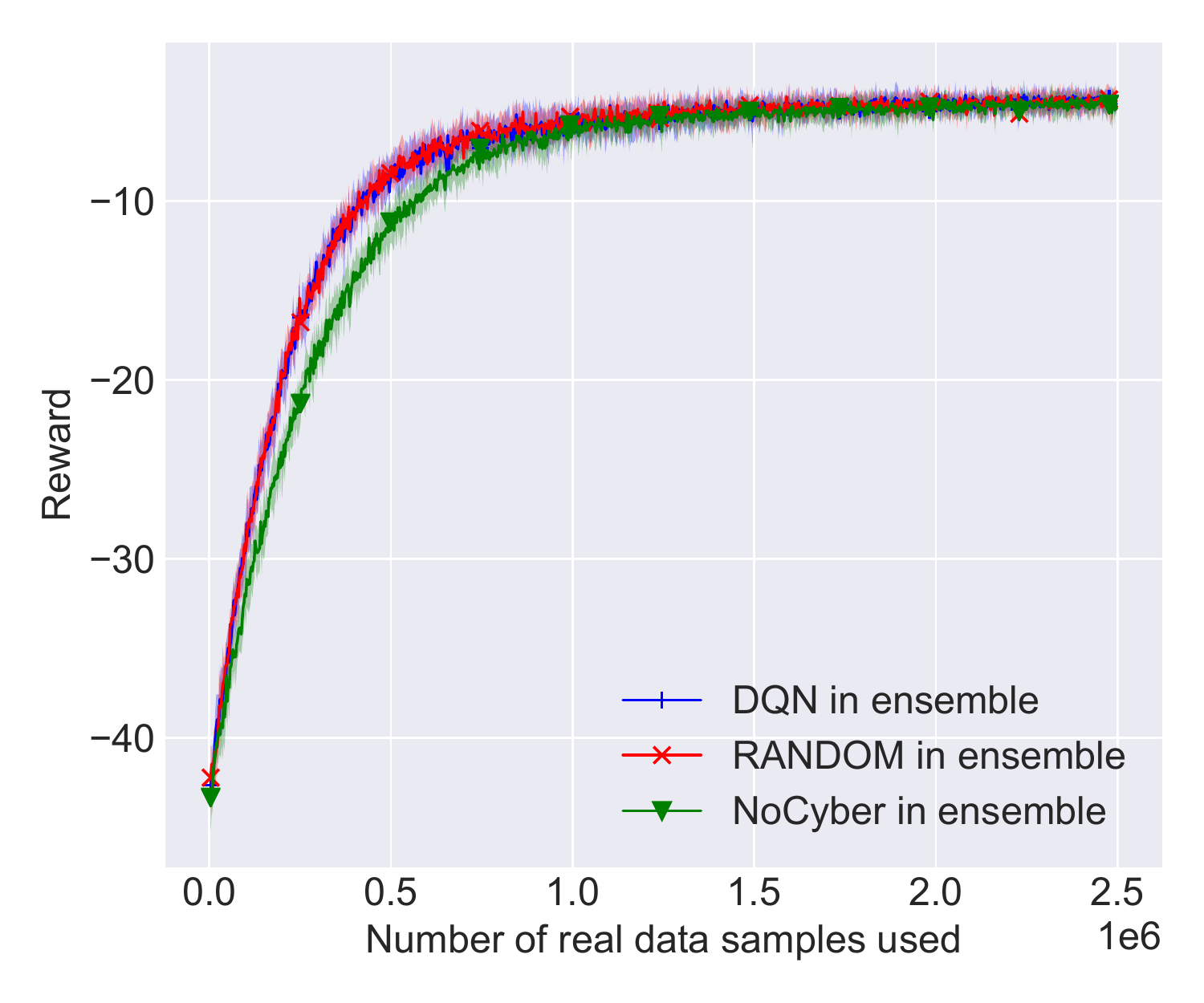}
}
\subfigure[Swimmer V1]{
   \includegraphics[width=0.3\textwidth] {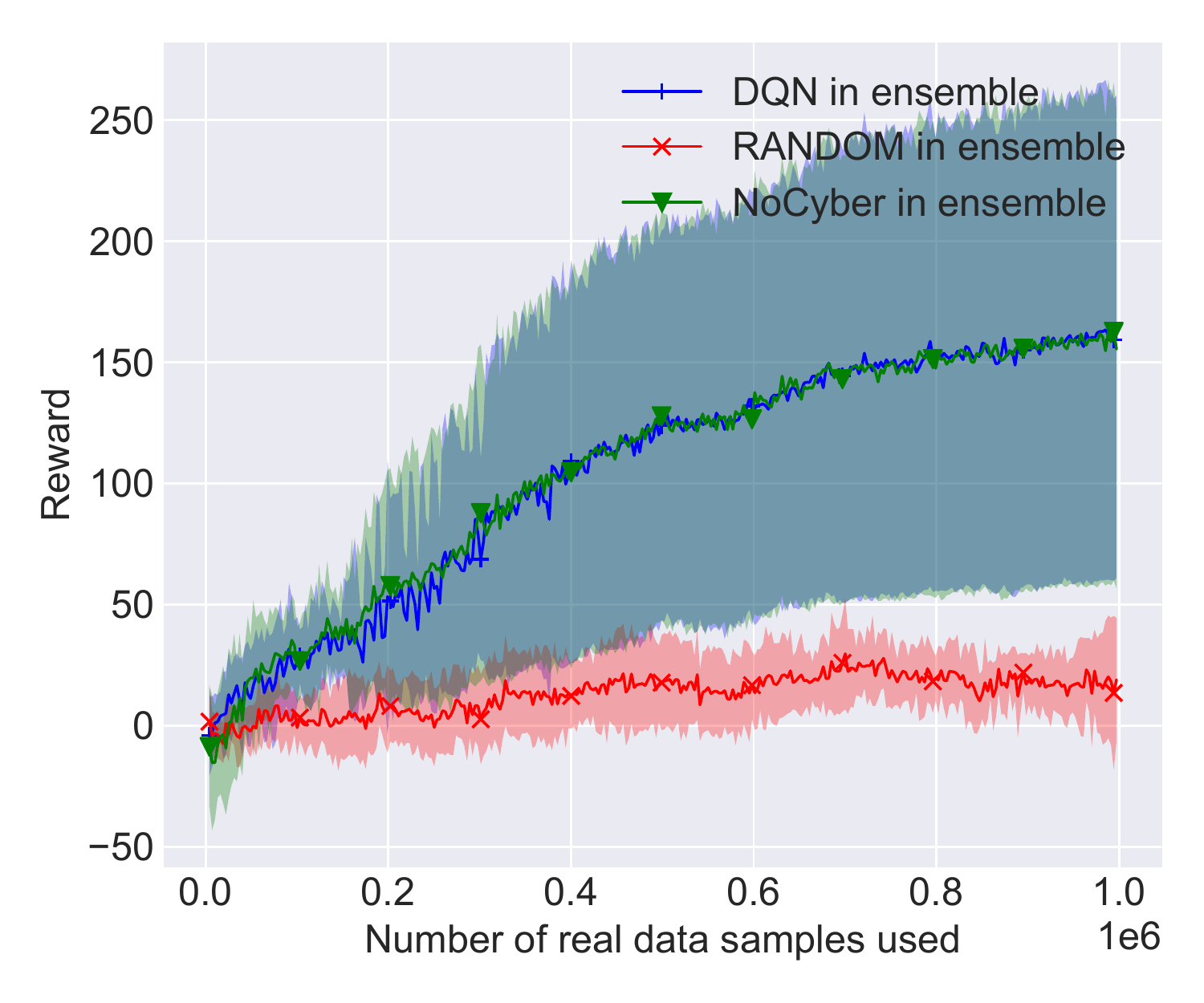}
}
\caption{Accumulative reward of different individual trainers of the ensemble trainer on (a) Mountain Car, (b) Reacher, and (c) Swimmer. Trainers' performance are tending to fuse except certain extremely under-performed trainers.}
\label{fig:ensemble_sep}
\end{figure*}

\subsection{Sensitivity Analysis on Various Trainer and TPE Designs}
\label{stn:diffdesign}
We compare the performances of different trainer and TPE designs  to study the performance sensitivity against the implementation variations. In addition to previously mentioned DQN, DQN-5 actions, and DQN-large memory, we also test DQN trainers with two different TPE state designs, as also listed in Table \ref{tbl:trainer_conf}. DQN-TPE V1 adopts the last average sampling reward of the target controller as the state of TPE; DQN-TPE V2 adopts the ratio (a value in the range of [0,1]) of the real samples used to the predefined maximum number of real samples as the state of TPE. Table \ref{tbl:variants_test} presents the accumulative rewards for five test cases: Pendulum, Mountain Car, Reacher, Half Cheetah, and Swimmer.

\begin{itemize}
 \item For Mountain Car, Reacher and Half Cheetah, DQN-5 actions, DQN-larger memory, DQN-TPE V1 and DQN-TPE V2 consistently outperform DQN. This indicates that for some applications, the intelligent trainer that uses more action selections, larger memory, or more informative state representation can achieve better performance. The results hint that a smart design of trainer or TPE can compensate the situation of lack of training data.
 \item For Swimmer, we observe that none of the tested variants of DQN or TPE can achieve satisfying performance. This is due to the fact that even a very small amount of cyber data can cause the target controller to be trapped in a local minimum that cannot be recovered.
\end{itemize}

\begin{figure}[]
\centering
   \includegraphics[width=0.6\columnwidth] {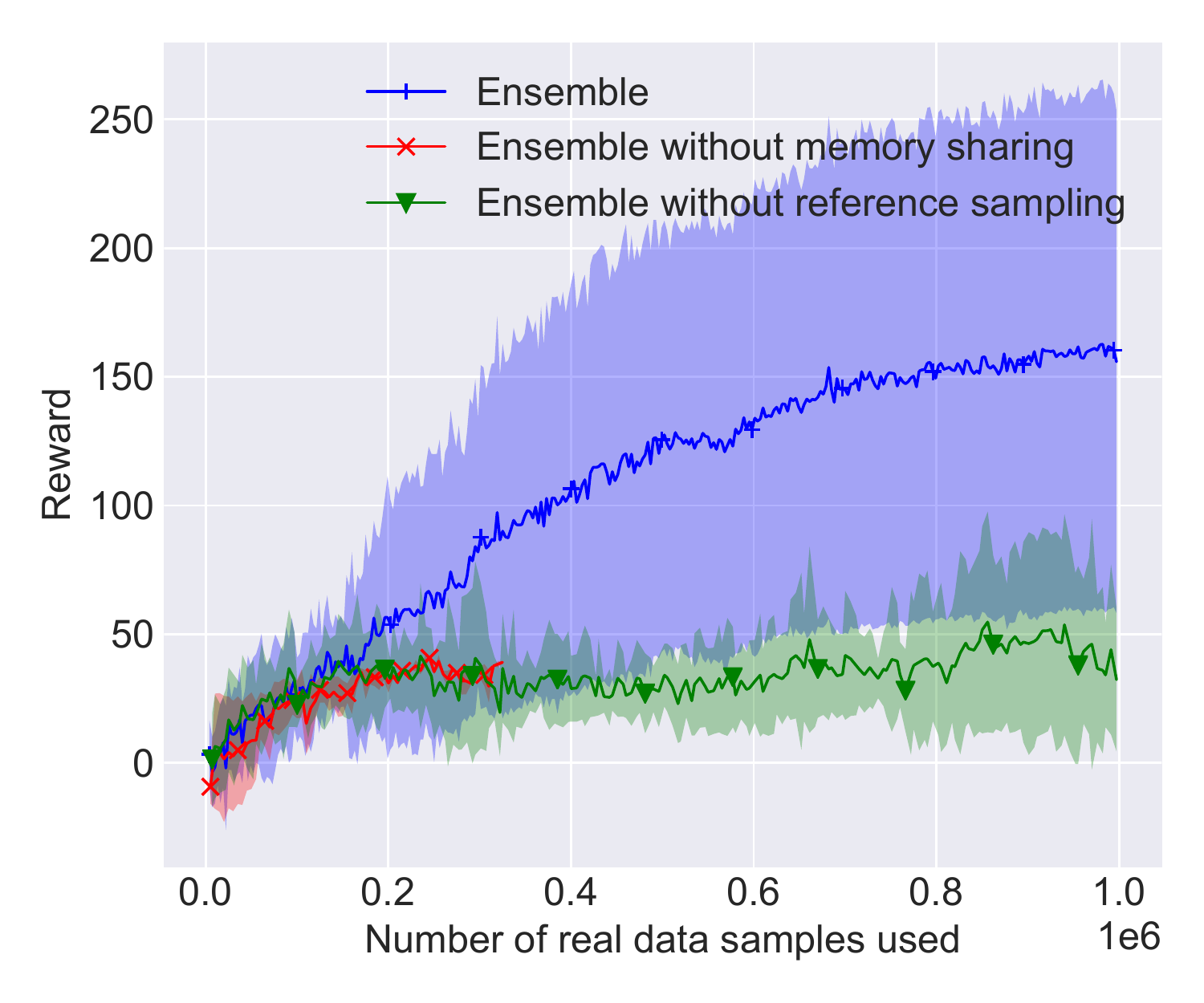}
\caption{Accumulative rewards of ensemble trainer and its two variants: without memory sharing and without reference sampling for Swimmer. The proposed ensemble design shows significant better performance.}
\label{fig:swimmer_ens_study}
\end{figure}

\subsection{Mitigating Action Correlation with Multi-head Ensemble Trainer}
As discussed in Section \ref{stn:ensTrainer}, the purpose of constructing an ensemble trainer is to overcome the action correlation problem in uni-head trainer. In this subsection, we provide evidence of the virtue of the ensemble design by comparing its performance with uni-head trainers. The ensemble trainer comprises a DQN trainer (with TPE state design V2), a Random Trainer, and a NoCyber trainer. Following the design in Section \ref{stn:ensTrainer}, these three trainers jointly sample and train three independent target controllers. The target controller of the best trainer will be used in the test. For the step threshold $C$ in weight transfer, it should be set to a TPE step count that a just sufficient number of trajectories (at least one episode) has been sampled. For such reason we set $C=3$ for all tasks except Mountain Car. For Mountain Car task, as in each TPE step, only one real sample is generated which is far from enough to evaluate the performance. We set to $C=100$ for this task. The upper and lower bounds $\phi_{max}$ and $\phi_{min}$ are estimated in the experiments, we found that $\phi_{max}=0.7$ and $\phi_{min}=0.5$ work well for all cases.

The results, as presented in Fig. \ref{fig:ensemble_results}, show that the ensemble trainer achieves overall good performance even in the cases the uni-head trainer fails. For the tasks of Pendulum, Mountain Car and Reacher, the ensemble trainer performs almost as well as the DQN or Random trainer. For the tasks of Swimmer and Half Cheetah, the ensemble trainer performs as well as the NoCyber trainer, even though the learning process makes it learn slower in the Half Cheetah task. With the proposed ensemble trainer, we are more likely to achieve sampling cost saving in practice as we it is hard to predict which kind of algorithm variant will deliver the best performance in advance. We compute the expected saving in Table \ref{tbl:saving} with the ensemble trainer when assuming the baseline sampling cost is the average cost of the three uni-head trainers NoCyber, DQN trainer and Random trainer. Note that for tasks Mountain Car, Half Cheetah and Swimmer, the uni-head trainer may fail to achieve the predefined performance target, in this case we set the cost as the maximum number of samples we tried in the experiment. That means the expected saving is actually larger than the number shown in Table \ref{tbl:saving}.

\begin{table}[]
\caption{Sampling saving to achieve certain predefined performance of the ensemble trainer. The baseline cost is the expected cost of the three algorithms NoCyber, Random trainer and DQN trainer.}
\label{tbl:saving}
\resizebox{\columnwidth}{!}{%
\begin{tabular}{c|c|c|c|c|c}
\hline
 & Pendulum & Mountain Car & Reacher & Half Cheetah & Swimmer \\ \hline
Target reward & -500 & 75 & -10 & 2500 & 100 \\ \hline
Samples saving & 26\% & 36\% & 2\% & 38\% & 56\% \\ \hline
\end{tabular}%
}
\end{table}

In Fig. \ref{fig:ensemble_results} (f), we observe that the action $a_2$ taken by the DQN varies significantly from the uni-head case. For Swimmer case, the action $a_2$ gradually converges to one which allows better performance. For Reacher case, we observe a phase transition in the middle, during which it changes from preferring fewer cyber data to more cyber data. This proves that when and how many cyber data should be utilized may be related to the training progress. For the Mountain Car task, we observe that it quickly converges to favor more cyber data which is helpful in this task. This proves that the proposed ensemble trainer can assess the control actions better than the uni-head trainer.

In Fig. \ref{fig:ensemble_sep}, we show the interactions of trainers in the ensemble by presenting individual results of the constituent trainers: DQN in ensemble, RANDOM in ensemble, and NoCyber in ensemble, for the tasks of Mountain Car, Reacher, and Swimmer (In the following of this paragraph, we omit the term of ``in ensemble'' for the sake of brevity). In all three cases, we can observe that within the ensemble, the original good trainer (uni-head) still performs very good. For example, for the Mountain Car task, the Random trainer performs almost as good as the uni-head Random trainer. For task Swimmer, the DQN trainer can now perform as good as the NoCyber trainer, which proves that the weight transfer process is working as expected.

To further examine the effect of memory sharing and reference sampling, in Fig. \ref{fig:swimmer_ens_study} we compare the performance of three different ensemble designs, for the task of Swimmer. All of them comprise the same three trainers: DQN, Random, and NoCyber, but differ in the incorporated schemes: ensemble trainer (with memory sharing and reference sampling); ensemble trainer without memory sharing (with reference sampling); ensemble trainer without reference sampling (with memory sharing). All these variants are with weight transfer. The results show that, without memory sharing, the ensemble performance degrades. This is because each of the three intelligent trainers uses only one-third of the original data samples (which is why the curve stops at 1/3 of the others in the x-axis). Without reference sampling, the ensemble performs very similar to the DQN trainer (Fig. \ref{fig:results_5case}). This is because without reference sampling, most of the real data samples are from underperformed target controllers of DQN and Random trainers. The data from underperformed target controllers deteriorates the learning process of the NoCyber trainer. The results indicate that memory sharing and reference sampling are essential for ensemble trainer.

\section{Related Works}
\label{stn:related}

To build intelligent agents that can learn to accomplish various control tasks, researchers have been actively studying reinforcement learning for decades, such as \cite{barto1983neuronlike,lewis2013reinforcement,liu2014policy,luo2015off,liu2015reinforcement}.
 With recent advancement of deep learning, DRL \cite{mnih2013playing} has demonstrated its strength in various applications. For example, in \cite{deng2017deep} a DRL agent is proposed to solve financial trading tasks;
in \cite{pan2017biomimetic} a neural RL agent is trained to mimic the human motor skill learning; in \cite{song2017off} an off-policy RL method is proposed to solve nonlinear and nonzero-sum games. Our research is particularly focused on model-based RL which can be utilized to reduce the sampling cost of RL, and we propose an AutoML method. In the following, we briefly review the recent development of model-based RL and the AutoML studies.

\subsection{Model-based RL}
Despite the significant performance improvement, the high sampling cost necessitated by RL has become a significant issue in practice. To address this issue, MBRL is introduced to learn the system dynamics model, so as to reduce the data collection and sampling cost. In \cite{deisenroth2011learning} the authors provided a MBRL for a robot controller that samples from both real physical environment and learned cyber emulator. In \cite{levine2015learning} the authors adapted a model, trained previously for other tasks, to train the controller for a new but similar task. This approach combines prior knowledge and the online adaptation of dynamic model, thus achieves better performance. In these approaches, the number of samples taken from the cyber environment to train the target controller is either predetermined or can only be adjusted manually, resulting in both sampling inefficiency and additional algorithm tuning cost. In \cite{kalweit2017uncertainty} the authors proposed a model-assisted bootstrapped DDPG algorithm, which uses a variance ratio computed from the multiple heads of the critic network to decide whether the cyber data can be used or not. The method relies on the bootstrapped DQN design, which is not suitable to other cases.

Instead of treating the cyber model as a data source for training, some approaches use cyber model to conduct pre-trial tree searches in applications, for which selecting the right action is highly critical \cite{guo2014deep} \cite{weber2017imagination}. The cyber model can prevent selecting unfavorable actions and thus accelerates the learning of the optimal policy. In \cite{pascanu2017learning}, the authors introduced a planning agent and a manager who decides whether to sample from the cyber engine or to take actions to minimize the training cost. Both approaches focuses on the tree search in action selection which is different to our design that we aim to select the proper data source in sampling.
Some recent works investigate integrating model-based and model-free approaches in RL. In \cite{baldi2015model} the authors combined model-based and model-free approaches for Building Optimization and Control (BOC), where a simulator is used to train the agent, while a real-world test-bed is used to evaluate the agent's performance. In \cite{nagabandi2017neural} the model-based DRL is used to train a controller agent. The agent is then used to provide weight initialization for a model-free DRL approach, so as to reduce the training cost. Different to this approach, we focus on directly sample from the model to reduce sampling cost in the real environment.

\subsection{AutoML}
The method proposed in this paper is a typical AutoML \cite{guyon2016brief} solution. AutoML aims to develop an algorithm that can automatically train a high performance machine learning model without human intervention, such as hyper-parameter tuning, model selection etc. AutoML has been proposed to solve various specific training tasks such as model compression for mobile device \cite{he2018amc}, transfer learning \cite{wong2018transfer}, general neural network training \cite{kim2017nemo}.

Note that most AutoML solutions are proposed to solve supervised learning cases, in which the dataset is usually pre-acquired. In our case, as the data will be collected by the target controller to train, it actually demands an AutoML solution more than a general supervised learning case.

\section{Conclusion}
\label{stn:summary}
In this paper we propose an intelligent trainer for online model training and sampling settings learning for MBRL algorithm. The proposed approach treats the training process of MBRL as the target system to optimize, and use a trainer that monitors and optimizes the sampling and training process in MBRL. The proposed trainer solution can be used in practical applications to reduce the sampling cost while achieve close-to-optimal performance.

For the future work, the proposed trainer framework can be further improved by adding more control actions to ease algorithm tuning cost. An even more advanced design is to use one trainer to train different DRL controllers for multiple tasks, which can learn the common knowledge shared by different DRL algorithms for these tasks.

\balance
\bibliographystyle{IEEEtran}
\bibliography{IEEEabrv,CFD_cali}

\end{document}